%% file: hpd_arxiv.tex
\documentclass{article}

\usepackage[T1]{fontenc}
\usepackage{hpd_arxiv,times}

\input{math_commands.tex}

\usepackage{amsmath}
\usepackage{amssymb}
\usepackage{amsfonts}
\usepackage{booktabs}
\usepackage{array}
\usepackage{graphicx}
\usepackage{microtype}
\usepackage{natbib}
\usepackage{placeins}
\usepackage{float}
\usepackage{needspace}
\usepackage{hyperref}
\usepackage{xurl}
\usepackage{xcolor}
\usepackage{pgfplots}
\usepackage{wrapfig}
\usepgfplotslibrary{groupplots}
\pgfplotsset{compat=1.18}
\newlength{\metriccolumnwidth}

\newcommand{\hpd}{\textsc{HPD}}
\newcommand{\replayaction}{\textsc{Harness-Replay}}
\newcommand{\injected}{\textsc{Injected}}
\newcommand{\runtime}{\textsc{Runtime}}
\newcommand{\autoprobe}{\textsc{AutoProbe}}
\newcommand{\githubicon}{\raisebox{-0.15\height}{\includegraphics[height=1.0em]{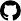}}}
\newcommand{\hficon}{\raisebox{-0.15\height}{\includegraphics[height=1.0em]{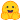}}}

\title{Harness Policy Distillation: Aligning Agent Policies with Runtime Semantics}
\author{%
  \textbf{Junze Zhu}\textsuperscript{1,2}\quad
  \textbf{Weihao Chen}\textsuperscript{1,2}\quad
  \textbf{Zhiyang Zhou}\textsuperscript{1,2}\quad
  \textbf{Zhen Wu}\textsuperscript{1,2}\thanks{Corresponding author.}\quad
  \textbf{Xinyu Dai}\textsuperscript{1,2}\\[6pt]
  {\fontsize{9}{11}\selectfont\textsuperscript{1}National Key Laboratory for Novel Software Technology, Nanjing University, China}\\
  {\fontsize{9}{11}\selectfont\textsuperscript{2}School of Artificial Intelligence, Nanjing University, China}\\
  {\fontsize{8}{10}\selectfont\texttt{\{zhujz, chenwh, zhouzy\}@smail.nju.edu.cn}}\\[-1pt]
  {\fontsize{8}{10}\selectfont\texttt{\{wuz, daixinyu\}@nju.edu.cn}}\\[6pt]
  {\small\href{https://github.com/HuiCir/HPD}{\githubicon\enspace Code}\quad\textbar\quad
  \href{https://huggingface.co/Migraine/Pi-Qwen3-8B}{\hficon\enspace Model}}
}
\hypersetup{%
  colorlinks=true, linkcolor=black, citecolor=black, urlcolor=black,
  pdftitle={Harness Policy Distillation: Aligning Agent Policies with Runtime Semantics},
  pdfauthor={Junze Zhu, Weihao Chen, Zhiyang Zhou, Zhen Wu, Xinyu Dai},
  pdfsubject={Public research preprint},
  pdfkeywords={language agents, harness policy distillation, runtime semantics, control-state supervision, tool use}
}
\date{}

\begin{document}
\maketitle
\raggedbottom

\begin{abstract}
Deploying a general-purpose model as an agent means running it inside a
product-specific runtime harness that decides which actions are executable, how
state advances, and when the loop may terminate. 
Ordinary trajectory supervision shows which action was taken but leaves the runtime semantics behind that action implicit. 
We introduce Harness Policy
Distillation (\hpd), which makes these runtime control semantics an explicit training target alongside actions to align the harness policy. 
HPD replays reference trajectories through the harness to recover the runtime semantics that are implicit in ordinary trajectory data, and represents them as control documents at each model-call boundary. These documents provide privileged information for a teacher to generate actions, while harness verification ensures that the resulting targets are executable under the runtime contract. We jointly trains a shared model on the runtime semantics recovered from control-state privileged information and native history, thereby internalizing these semantics into the model policy. At deployment, the trained model uses the original harness interface without requiring the control document or any additional supplement. 
On TRAJECT-Bench with the Pi harness,
Qwen3-8B trained with
\hpd{} reaches 22.98\% autonomous task success and 49.23\% runtime compliance,
against 3.09\% and 19.90\% for the unadapted model. Harness-verified teacher actions substantially improve the native deployment policy, while joint control-state supervision provides further gains in task completion and runtime compliance. HPD turns executable runtime semantics into dense supervision for learning harness-compatible actions at each decision boundary.
\end{abstract}

\section{Introduction}

A general-purpose model becomes a deployed agent only inside a product-specific
runtime harness. The harness serializes visible history, exposes an approved
action interface, enforces permissions and workflow stages, inserts tool
observations, schedules subsequent model calls, and defines valid termination
\citep{yang2024sweagent,wang2025openhands}. For an organization adapting a model
to its own product, the relevant target is therefore not task competence or tool
syntax alone, but a harness-bound policy whose actions remain executable under
that product's runtime contract.

Most post-training objectives supervise behavior without separately specifying the runtime control state on which that behavior depends. Demonstration and trajectory training maximize the likelihood of reference actions conditioned on observed histories, leaving the control state implicit in the mapping from history to action. Direct Preference Optimization (DPO) increases the relative likelihood of preferred outputs \citep{rafailov2023direct}; Reinforcement learning with verifiable rewards (RLVR) optimizes expected return from executable checks \citep{shao2024deepseekmath,deepseekai2025r1}, leaving the policy to infer reward-relevant control state through credit assignment. On-policy distillation (OPD) matches teacher behavior at student-generated prefixes \citep{agarwal2024onpolicy}, improving the coverage of states at which supervision is supplied while leaving the teacher's action distribution as the direct target. These objectives can learn runtime-sensitive behavior, but their standard forms leave the underlying control state to be inferred through action, preference, reward, or teacher supervision.

Within an agent, the model selects actions while the harness governs
their execution. We distinguish two sources of policy mismatch
between them. Supervision mismatch arises when the histories
used for supervision differ from those generated by the agent's own
rollout: a model may fit reference actions yet struggle to continue
once its own actions lead to histories poorly covered during training
\citep{ross2011reduction,bengio2015scheduled}.
Runtime-state mismatch arises when the model policy fails to
reflect the harness's current execution requirements. Even on a valid
reference history, action labels leave implicit which obligations
remain and whether termination is allowed; the model may therefore
repeat completed work or stop while required work is still pending. The two problems interact through the closed loop: each action changes the
successor state and hence the context of the next decision. Improving the
conditional policy at replayed boundaries is therefore a concrete training
target, while autonomous task completion measures its effect across an entire
execution. Because each action changes the next runtime state and model context,
a local error can affect later decisions.

The harness provides a complementary source of supervision: its execution history and runtime bookkeeping identify the obligations and lifecycle conditions that govern the next action. For example, tool availability alone does not indicate whether the corresponding obligation has already been completed. Learning the next action from a reference trajectory leaves this distinction implicit; explicitly supervising the boundary's control state makes it a prediction. We therefore target this policy mismatch with explicit control-state and action supervision at replayed boundaries, and evaluate its effect through autonomous task completion.

\begin{figure}[!t]
\centering
\includegraphics[width=\linewidth]{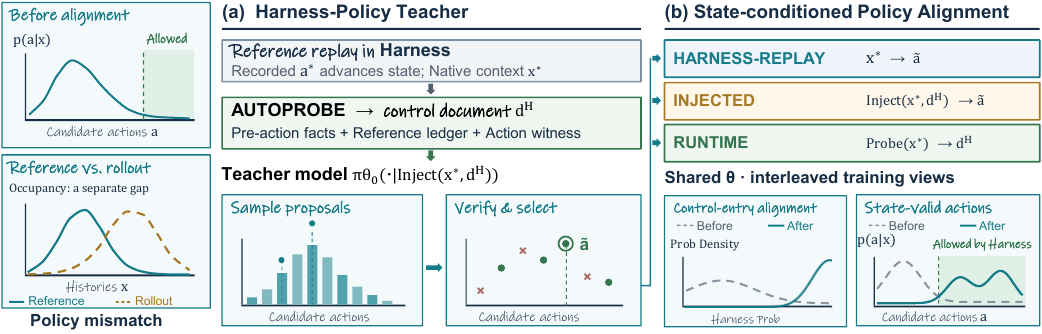}
\caption{Harness Policy Distillation. Left: agent model--harness policy mismatches.
(a) Recorded actions advance replay; \autoprobe{} compiles control documents;
teacher-model proposals are verified and ranked by reference similarity.
(b) Three views
train one shared model for native action prediction, document-conditioned
action prediction, and document reconstruction.}
\label{fig:hpd-main}
\end{figure}

To make this connection learnable, we propose \textbf{Harness Policy
Distillation (HPD)}, which couples explicit control-state supervision
with harness-verified action targets to align the model policy with
the runtime semantics governing execution. At each replayed boundary,
HPD compiles runtime observations and reference-derived control
annotations into a privileged control document. This document guides
a sampling model to propose actions, while harness verification
restricts the training targets to candidates consistent with the
execution and reference-control requirements. Recorded actions
advance replay while selected candidates provide training targets,
preserving the reference histories on which the paired supervision
is defined.

To internalize these control semantics without requiring the document
during autonomous execution, HPD jointly trains three views of one
shared model. \textsc{Runtime} uses the compiled document as a target
for control-state prediction from native history; \textsc{Injected}
teaches action selection with these conditions made explicit.
\textsc{Harness-Replay} trains the same verified action from native
context alone, so action learning also matches the input available
at deployment. Together, these views connect explicit control
supervision to native action selection. The trained model acts through
the original harness without an external control document or an
additional model call (Figure~1).

\section{Related Work}
\label{sec:related-work}

\paragraph{Trajectory learning and policy distillation.}
Policies trained by behavior cloning run on histories induced by their own actions,
motivating interactive imitation and scheduled sampling
\citep{ross2011reduction,bengio2015scheduled}. Sequence models instead fit
state--action trajectories directly \citep{chen2021decision,janner2021trajectory},
while knowledge distillation and language-model post-training transfer teacher
sequences, demonstrations, or preferences
\citep{hinton2015distilling,kim2016sequence,ouyang2022training,
rafailov2023direct}. On-policy distillation and agentic RL move supervision or
reward toward student-generated prefixes
\citep{agarwal2024onpolicy,shao2024deepseekmath,deepseekai2025r1}. These methods
primarily change the action target, reward, or visited occupancy; \hpd{} instead
adds runtime facts and reference-derived control labels as supervised targets
at replayed decision boundaries.

\paragraph{Privileged and process supervision.}
Learning using privileged information introduces training-time variables that
need not be available at inference
\citep{vapnik2009new,lopezpaz2016unifying}. In self-distillation, privileged
context lets a model supervise its counterpart without that context
\citep{zhao2026selfdistilled}. $\pi$-Distill jointly trains a privileged-information-conditioned teacher
and a student without that information using shared model parameters
in multi-turn agentic environments \citep{penaloza2026privileged}. Process supervision identifies where a multistep solution becomes invalid \citep{uesato2022solving,lightman2024verify}. Training with Harnesses
uses a harness-augmented current model as an on-policy self-teacher to
internalize reasoning workflows \citep{zhao2026trainingharnesses}. HPD pairs
replay-derived control documents with actions selected by harness verification
at the same decision boundary.

\paragraph{Tool agents and runtime harnesses.}
ReAct and tool-learning methods interleave language-model generation with
environment actions or construct tool-use supervision
\citep{yao2023react,schick2023toolformer,yang2023gpt4tools,qin2024toolllm}.
Interactive benchmarks make stateful execution central
\citep{shridhar2021alfworld,yao2022webshop,zhou2024webarena,
liu2024agentbench,mialon2024gaia,jimenez2024swebench}. SWE-agent and OpenHands
further show that the agent interface, event stream, sandbox, and stopping rules
are part of the deployed system \citep{yang2024sweagent,wang2025openhands}.
Evaluation work separates final task completion from intermediate progress and
policy compliance \citep{ma2024agentboard,yao2025taubench,lu2025toolsandbox}.
\hpd{} uses this runtime lifecycle as a supervised object rather than treating
it only as an external evaluator.

\section{Methodology}
\label{sec:method}
\subsection{Closed-loop execution under a runtime harness}
\label{sec:closed-loop}

Let $q$ denote a task instance, comprising the request and execution resources,
and $H$ the target runtime harness. Its native initialization produces the
initial state $s_0$ from that instance. The state space $\mathcal S_H$ contains
complete runtime configurations: execution history, tool interfaces, and
the control fields maintained by $H$, including its permissions, scheduling
state, obligation bookkeeping, and remaining budget.
For replay, the fixture additionally supplies recorded tool-observation queues;
their consumption positions are part of the state.
Valid stopping states form $\mathcal S_H^{\mathrm{term}}\subseteq\mathcal S_H$.
Add an absorbing failure state $\bot\notin\mathcal S_H$ for fatal contract
violations or budget exhaustion, giving
$\bar{\mathcal S}_H=\mathcal S_H\cup\{\bot\}$. Assume finite execution and let
$B$ be the first index with $s_B\in\mathcal S_H^{\mathrm{term}}\cup\{\bot\}$.
For $b=0,\ldots,B-1$, $s_b$ is the complete state immediately before model call
$b$, and $x_b$ is its serialized model-visible context.

Let $\mathcal X_H$ be the space of native contexts and $\mathcal A$ the space
of complete model outputs interpreted as actions, including tool calls and
final responses. The harness exposes a serializer
$\Omega_H:\mathcal S_H\to\mathcal X_H$ and a transition kernel:
\begin{equation}
x_b=\Omega_H(s_b),
\qquad
T_H:\bar{\mathcal S}_H\times\mathcal A
\rightarrow\Delta(\bar{\mathcal S}_H),
\label{eq:harness-operators}
\end{equation}
where $\Delta$ denotes probability distributions over its argument space.
$T_H(\cdot\mid s,a)$ is the successor law induced by executing action $a$ from
state $s$ through the harness's parsing, permission checks, tool execution,
observation insertion, scheduling, and stopping code. The successor is the
next pre-call state or the terminal/failure outcome. Terminal and failure
states are absorbing; recoverable errors become observations in nonterminal
states. Thus $T_H$ is supplied by the runtime implementation, while the
\emph{harness policy} denotes its state-dependent execution and stopping rules.

For a nonterminal state $s$, the immediately admissible action set is
\begin{equation}
\mathcal A_H(s)=
\left\{a\in\mathcal A:T_H(\bot\mid s,a)=0\right\}.
\label{eq:admissible-actions}
\end{equation}
For deterministic replay, admissible actions reach a successor in
$\mathcal S_H$. Reference-path agreement evaluates a sequence of decisions,
while task success evaluates completed obligations, the answer, and termination.
The model $\pi_\theta$ assigns sequence probabilities given a textual prompt,
with trainable parameters $\theta$. Its harness action policy is
$\pi_\theta(a\mid x)$ for $x\in\mathcal X_H$. It generates
the closed loop from the harness-initialized state $s_0$:
\begin{equation}
x_b=\Omega_H(s_b),\qquad
a_b\sim\pi_\theta(\cdot\mid x_b),\qquad
s_{b+1}\sim T_H(\cdot\mid s_b,a_b).
\label{eq:closed-loop-step}
\end{equation}
For a trajectory $\tau=(s_0,x_0,a_0,\ldots,s_{B-1},x_{B-1},a_{B-1},s_B)$
with $x_b=\Omega_H(s_b)$, conditioning on the task and initialization gives
\begin{equation}
p_{\theta,H}(\tau\mid q,s_0)=\prod_{b=0}^{B-1}
\pi_\theta(a_b\mid\Omega_H(s_b))
T_H(s_{b+1}\mid s_b,a_b).
\label{eq:rollout-distribution}
\end{equation}
The executable agent is thus the process jointly induced by $\pi_\theta$,
$\Omega_H$, and $T_H$. Each action determines a successor context, connecting
boundary-level learning to closed-loop task completion.

\subsection{Runtime semantic alignment gap}
\label{sec:pd-axes}

Policy distillation specifies both a context distribution $\mu$ and a teacher
conditional $\pi^{\mathrm T}(\cdot\mid x)$ over actions. Here
$D_{\mathrm{KL}}$ denotes forward Kullback--Leibler divergence, and
$\mathbb E$ denotes averaging under the indicated distribution:
\begin{equation}
\mathcal L_{\mathrm{PD}}(\theta;\mu,\pi^{\mathrm T})
=\mathbb E_{x\sim\mu}
\left[D_{\mathrm{KL}}\!\left(
\pi^{\mathrm T}(\cdot\mid x)\,\Vert\,
\pi_\theta(\cdot\mid x)\right)\right].
\label{eq:generic-pd}
\end{equation}
For a fixed task collection, $\mu_H^\star$ is the empirical context
distribution counting each replayed reference boundary once;
$\mu_{\theta,H}$ normalizes the expected boundary counts under free rollout
on that collection.
Their discrepancy is the supervision mismatch; on-policy distillation moves
supervision toward $\mu_{\theta,H}$.

Runtime-state mismatch concerns the conditional target at a given boundary.
An action label specifies what was emitted; runtime control labels additionally
identify lifecycle stage, remaining obligations, ready calls, and termination
conditions. HPD supervises these labels together with a verified action at
reference-history boundaries. Verification yields retained records; uniform sampling of these records
defines the context distribution $\mu_H^{\mathrm{rep}}$. This filtering
can change the relative weights of reference histories. Training fits the paired
control and action targets there, and free rollout evaluates the resulting
policy on its own successive decisions.

\subsection{Harness Policy Distillation}
\label{sec:hpd}

HPD addresses this conditional-target gap by constructing paired control
and action supervision at replayed reference boundaries
(Figure~\ref{fig:hpd-main}).
Reference actions advance the harness, while runtime observations and
reference annotations are compiled into a privileged control document at
each boundary. The frozen base model conditions on this document to propose
actions, from which verification and near-reference selection produce
the training target. The resulting context--document--action triples
train one shared model through native-context action prediction,
document-conditioned action prediction, and control-document reconstruction.
At deployment, the model predicts actions directly from the native context.

\paragraph{Verified replay and control-document construction.}
Let $\tau^\star$ be a recorded task trajectory containing actions, tool
observations, and reference stages; $a_b^\star$ is its action aligned to the
$b$-th harness invocation. Superscript $\star$ marks reference replay, and
superscript $H$ marks harness-specific annotations. Starting from native
initialization $s_0^\star$, replay submits each $a_b^\star$ to $H$, supplies
recorded observations from per-tool queues, and captures the resulting
$s_{b+1}^\star$ at the next boundary. The pre-action observation adapter
$\operatorname{Observe}_H$ reads accessible runtime fields into snapshot
$o_b^H$; the deterministic compiler $\Phi_H$ maps that snapshot and the
reference annotations to a textual control document $d_b^H$.
Each retained boundary satisfies
\begin{equation}
\begin{aligned}
x_b^\star
&=\Omega_H(s_b^\star),\qquad
o_b^H=\operatorname{Observe}_H(s_b^\star),\\
d_b^H
&=\Phi_H(o_b^H;\tau^\star,a_b^\star),\\
a_b^\star
&\in\mathcal A_H(s_b^\star),\qquad
T_H(s_{b+1}^\star\mid s_b^\star,a_b^\star)>0.
\end{aligned}
\label{eq:verified-replay}
\end{equation}
The transition condition checks the successor actually produced by replay;
for deterministic replay its probability is one. \autoprobe{} comprises
$\operatorname{Observe}_H$ and $\Phi_H$. The document $d_b^H$ contains
observed runtime facts, reference-stage progress, and witness-derived labels
for the next control, ready frontier, and final permission. The latter fields
provide privileged action information during training; field provenance and
harness-specific adapters are detailed in Appendix~\ref{app:replay-details}.

For illustration, specify a contract requiring
\texttt{lookup(item\_id=42)} before a final response, and a replay queue
containing \texttt{value=7}. Initially lookup is pending and finalization is
blocked. Executing the recorded call consumes that observation, completes
the lookup obligation, and enables finalization. The documents label these
pre-action states as \texttt{lookup=pending; final=blocked} and
\texttt{lookup=complete; final=ready}. These transitions follow from the
specified contract.

\paragraph{Harness-verified privileged teacher.}
Let $\theta_0$ be the frozen base-model parameters. The prompt constructor
$\operatorname{Inject}_H(x,d)$ supplies document $d$ alongside native context
$x$. The model samples $K$ candidate actions $\tilde a_b^{(k)}$ per attempt,
where $k$ indexes candidates; the accepted set is $\mathcal A_b^{\mathrm{ver}}$
and its selected action is $\tilde a_b$:
\begin{equation}
\begin{aligned}
\tilde a_b^{(k)}&\sim\pi_{\theta_0}\!\left(
\cdot\mid\operatorname{Inject}_H(x_b^\star,d_b^H)\right),
\qquad k\in\{1,\ldots,K\},\\
\mathcal A_b^{\mathrm{ver}}&=\left\{
\tilde a_b^{(k)}:\;V_{H,b}(s_b^\star,\tilde a_b^{(k)})=1
\right\},\\
\tilde a_b&\in\mathcal A_b^{\mathrm{ver}}.
\end{aligned}
\label{eq:teacher-sampling}
\end{equation}
$V_{H,b}(s,a)\in\{0,1\}$ checks admissibility, reference-stage consistency,
and exact agreement with the recorded control decision, tool names, and complete
normalized tool arguments. We select one verified candidate by cosine
similarity to the reference over its remaining fields. The recorded action
$a_b^\star$ advances replay, while the selected
$\tilde a_b$ supplies the training target. Appendix~\ref{app:teacher-construction}
details parameter replay, candidate selection, corrective resampling, and
reference-based continuation during offline construction. Each retained record is the triple
$(x_b^\star,d_b^H,\tilde a_b)$.

\paragraph{Three-view training and native deployment.}
Let $d$ range over control-document strings. The constructor
$\operatorname{Probe}_H(x)$ requests document prediction from native history
without supplying the target document. Superscripts T, I, and R denote the
three prompt contracts of one shared model:
\begin{equation}
\begin{aligned}
\pi_\theta^T(a\mid x)
&=\pi_\theta(a\mid x),\\
\pi_\theta^I(a\mid x,d)
&=\pi_\theta(a\mid\operatorname{Inject}_H(x,d)),\\
\pi_\theta^R(d\mid x)
&=\pi_\theta(d\mid\operatorname{Probe}_H(x)).
\end{aligned}
\label{eq:three-view-policies}
\end{equation}
\replayaction{} (T) predicts $\tilde a_b$ from the native context;
\injected{} (I) predicts the same action with $d_b^H$ supplied;
\runtime{} (R) predicts $d_b^H$ from native history. These views connect
history to control, control to action, and history directly to the deployed
action through shared parameters. The paired document--action records define
a joint teacher: R fits its document marginal, I its action conditional given
the document, and T its native-runtime action marginal. By the KL chain rule,
the R and I losses decompose joint-teacher matching; T directly supervises
the deployment policy. Let $\mathcal L_T$, $\mathcal L_I$, and $\mathcal L_R$
be these conditional teacher-to-student KL losses averaged over retained
records (Appendix~\ref{app:derivation}). Their weighted objective is
\begin{equation}
\mathcal L_{\mathrm{HPD}}
=\alpha_T\mathcal L_T+\alpha_I\mathcal L_I+\alpha_R\mathcal L_R,
\qquad
\alpha_T+\alpha_I+\alpha_R=1.
\label{eq:hpd-objective}
\end{equation}
Here $\alpha_v\geq0$ is the supervision frequency for view
$v\in\{T,I,R\}$.

\paragraph{Sequence-level distillation and implementation.}
For each view, let $c$ be its prompt from
Equation~\ref{eq:three-view-policies} and $y$ its stored completion:
$\tilde a_b$ for T/I or $d_b^H$ for R. The point mass $\delta_y$ assigns
probability one to $y$, including its completion boundary. With $|y|$ tokens,
token $y_j$, and target prefix $y_{<j}$, autoregressive factorization gives
\begin{equation}
D_{\mathrm{KL}}\!\left(\delta_y\,\Vert\,\pi_\theta(\cdot\mid c)\right)
=-\log\pi_\theta(y\mid c)
=-\sum_{j=1}^{|y|}\log\pi_\theta(y_j\mid c,y_{<j}).
\label{eq:hard-target-kl}
\end{equation}
Teacher-forced token prediction thus implements sequence-level forward KL
for each hard target \citep{kim2016sequence}. Averaging per-record NLLs
recovers each view's conditional teacher KL up to a constant teacher entropy;
Appendix~\ref{app:derivation} derives this aggregation for shared prompts.
The implemented loss $\ell_\theta(y,c)=-\log\pi_\theta(y\mid c)/|y|$
weights each sequence KL by inverse completion length, putting documents,
final responses, and tool-call waves on a per-token scale.
Training uses stored targets and cycles through view-specific batches. Deployment uses only $\pi_\theta^T(\cdot\mid x)$;
the full implemented objective is given in Appendix~\ref{app:derivation}.

\section{Experiments}
\label{sec:experiments}

\begingroup
\setlength{\textfloatsep}{9pt plus 2pt minus 2pt}
\setlength{\floatsep}{8pt plus 2pt minus 2pt}
\setlength{\intextsep}{7pt plus 2pt minus 2pt}
\setlength{\abovecaptionskip}{4pt}
\setlength{\belowcaptionskip}{0pt}

\subsection{Setup}

We evaluate Qwen3-8B \citep{yang2025qwen3} with LoRA adaptation
\citep{hu2022lora} on 583 held-out TRAJECT-Bench tasks
\citep{he2026trajectbench}. Table~\ref{tab:main} compares HPD with the base
model and supervised, preference, reinforcement-learning, and distillation
baselines. We report task success (TS), final-answer F1, gold-history action
accuracy (Action-Acc), reference-prefix success (Step-SR), runtime compliance
(RC), and exact-path success (Exact), all produced by \texttt{pi-runtime}.
We then test component and schedule effects, boundary behavior, adaptation
across models and harnesses, and transfer to a new task collection.
Appendices~\ref{app:experimental-details} and~\ref{app:metrics} specify the
training, baseline implementations, seed study, and metric definitions.

\subsection{Main comparison}
\label{sec:main}

\begin{table}[!htbp]
\centering
\small
\setlength{\tabcolsep}{2pt}
\settowidth{\metriccolumnwidth}{Action-Acc}
\caption{Qwen3-8B comparison on TRAJECT-Bench/Pi. Scores (\%) are from
\texttt{pi-runtime}; TS, RC, and Exact use 583 cases. HPD-T and HPD share
document-conditioned, verifier-selected teacher targets; HPD-T trains only
the Harness-Replay (T) view, while HPD adds I/R. Baselines are detailed in
Appendix~\ref{app:experimental-details}.}
\label{tab:main}
\begin{tabular}{@{}p{0.11\linewidth}p{\dimexpr\linewidth-0.11\linewidth-6\metriccolumnwidth-14\tabcolsep\relax}*{6}{>{\centering\arraybackslash}p{\metriccolumnwidth}}@{}}
\toprule
Method & Supervision source & TS & F1 & Step-SR & RC & Exact & Action-Acc \\
\midrule
Base & --- & 3.09 & 26.82 & 26.45 & 19.90 & 2.57 & 53.41 \\
SFT & Ground truth & 2.06 & 2.40 & 47.20 & 55.92 & 1.72 & 2.06 \\
DPO & Preference labels & 7.20 & 33.62 & 11.12 & 25.56 & 5.15 & \textbf{72.38} \\
GiGPO & Task reward & 4.46 & 29.65 & 26.17 & 21.61 & 3.43 & 54.63 \\
GRPO & Task reward & 3.43 & 27.59 & 25.79 & 20.41 & 2.74 & 54.35 \\
GSPO & Task reward & 7.72 & 37.61 & 13.46 & 28.64 & 6.69 & 54.30 \\
GDPO & Task reward & 4.63 & 35.04 & 19.07 & 23.50 & 3.77 & 54.11 \\
OPD & Teacher model & 15.61 & 41.04 & 57.76 & \textbf{66.21} & 14.07 & 59.91 \\
OPSD & Privileged info & 13.55 & 26.27 & 55.98 & 26.93 & 10.98 & 63.97 \\
OPHSD & Teacher + harness & 3.09 & 29.45 & 27.66 & 19.73 & 2.06 & 53.22 \\
TurnOPD & Teacher model & 9.43 & 29.02 & 50.70 & 44.77 & 6.52 & 59.77 \\
\midrule
HPD-T & Harness runtime & 20.58 & 39.96 & 59.25 & 43.74 & 15.61 & 61.40 \\
\textbf{HPD} & \textbf{Harness runtime} & \textbf{22.98} & \textbf{41.22} & \textbf{63.79} & 49.23 & \textbf{18.70} & 64.95 \\
\bottomrule
\end{tabular}
\end{table}

Table~\ref{tab:main} separates local ranking (Action-Acc), execution
(Step-SR and RC), answer overlap (F1), and completion (TS and Exact).
HPD-T reaches 20.58 TS, above the strongest baseline, OPD at 15.61.
Full HPD leads in TS, F1, Step-SR, and Exact: its 22.98 TS is 43 completed
tasks above OPD (7.38 points; 47.3\% relative).
The five-seed paired analysis supports both TS differences
(Appendix~\ref{app:statistical-reporting}).

\textbf{Reference supervision: SFT and DPO.}
SFT raises RC from 19.90 to 55.92 and Step-SR from 26.45 to 47.20, yet lowers
TS from 3.09 to 2.06 and F1 from 26.82 to 2.40. Fitting recorded gold
actions improves reference-path metrics here without preserving answer quality
or task completion under Pi deployment, illustrating the mismatch introduced
in the Introduction. DPO achieves the best
Action-Acc alongside 11.12 Step-SR and 7.20 TS. Its advantage is local
reference-action ranking; its rollout follows the reference prefix less often.
RC separately measures compliance with the execution contract. Together,
these metrics distinguish fitting the next reference action, maintaining its
reference path, and completing an executable task.

\textbf{Task rewards: GRPO and its variants.}
GiGPO and GRPO give modest TS gains (4.46 and 3.43), with Step-SR
(26.17 and 25.79) close to Base. GSPO leads the reward-based baselines in
TS and F1 (7.72 and 37.61); GDPO reaches 35.04 F1 but only 4.63 TS. Their
Step-SR values (13.46 and 19.07) show that higher answer overlap need not
sustain the reference path.

\textbf{Model teachers: self-distillation and OPD.}
OPHSD leaves TS at 3.09. OPSD reaches 13.55 TS and 55.98 Step-SR
with F1 near Base (26.27 versus 26.82), showing stronger execution without
higher answer overlap. OPD achieves the highest RC (66.21), with
57.76 Step-SR, 15.61 TS, and 14.07 Exact at 41.04 F1.
TurnOPD also improves reference-prefix execution and compliance over Base,
reaching 50.70 Step-SR, 44.77 RC, 9.43 TS, and 6.52 Exact. OPD has the
highest RC, while HPD's higher TS shows that compliance and completed tasks
rank differently.

\textbf{Harness-verified replay and runtime semantics.}
HPD-T uses the same document-conditioned, verifier-selected teacher as Full
HPD, but trains only the native replay-action view. Relative to OPD, it gains
4.97 TS points and 1.54 Exact points while its mean F1 is slightly lower
(39.96 versus 41.04). The verified-target and native-action pipeline thus
improves closed-loop completion. Adding I/R training views
raises TS by another 2.40 points, Step-SR by 4.54 points, and Exact by
3.09 points. The runtime semantics injection and control-document views contribute to this second gain.

\subsection{Cross-benchmark robustness study}
\label{sec:cross-benchmark}

\begin{wrapfigure}{r}{0.49\linewidth}
  \centering
  \vspace{-0.6\baselineskip}
  \includegraphics[width=\linewidth]{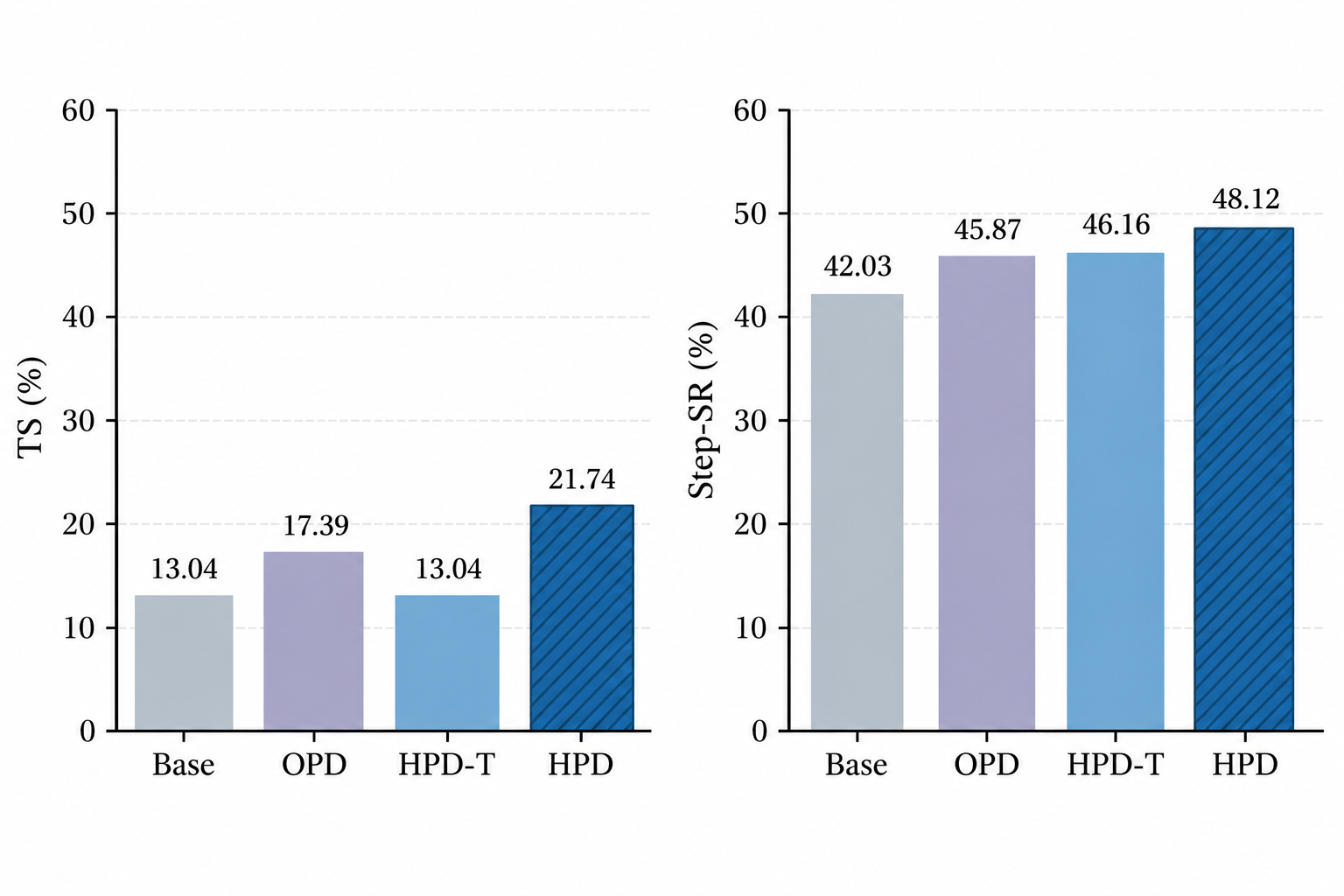}
  \caption{Zero-shot transfer to BFCL v4 from above checkpoints without specific training.}
  \label{fig:bfcl-main}
  \vspace{-0.4\baselineskip}
\end{wrapfigure}
We evaluate the TRAJECT-Bench-trained checkpoints on 92 BFCL v4 cases
\citep{patil2025bfcl} without BFCL-specific training. BFCL supplies new
tasks and default tool execution; the Pi agent, checkpoints, and model-side
settings remain fixed (Appendix~\ref{app:bfcl-ood}). Full HPD reaches
21.74 TS, and 48.12 Step-SR versus Base's 13.04, and
42.03. HPD-T reaches 13.04 TS, and 46.16 Step-SR: adding I/R
raises TS by 8.70 and Step-SR by 1.96 points. 

HPD-T improves answer overlap and prefix execution over Base but not task
completion. On these new tasks, Full HPD's completion gain is consistent
with I/R connecting replay actions to Pi's pending obligations and
termination beyond source-task patterns. When the test tasks extend beyond the range of trajectories encountered during training, the performance gains of HPD-T drop significantly; in contrast, HPD reinforces the alignment of the harness policy---rather than overfitting to a specific task distribution---by leveraging runtime semantic training (I/R), which accounts for its superior generalization capabilities.

\subsection{HPD components and training schedule}
\label{sec:ablation}

Table~\ref{tab:ablation} compares the replay-action view and
control-document views at matched exposure: each active view sees
20k training batches. Joint variants interleave view updates, whereas staged
variants apply 20k batches per view in consecutive phases. Appendix~\ref{app:step-accounting} explains the bookkeeping.

\begin{table}[!htbp]
\centering
\small
\setlength{\tabcolsep}{2pt}
\settowidth{\metriccolumnwidth}{Action-Acc}
\caption{Qwen3-8B HPD component and schedule ablation with 20k training
batches per active view. T is the replay-action view; I and R are the injected-
and runtime-document views. Arrows denote consecutive phases. Scores are
percentages from \texttt{pi-runtime}.}
\label{tab:ablation}
\begin{tabular}{@{}p{\dimexpr\linewidth-6\metriccolumnwidth-12\tabcolsep\relax}*{6}{>{\centering\arraybackslash}p{\metriccolumnwidth}}@{}}
\toprule
HPD variant & TS & F1 & Step-SR & RC & Exact & Action-Acc \\
\midrule
HPD-T & 20.58 & 39.96 & 59.25 & 43.74 & 15.61 & 61.40 \\
T$\rightarrow$I (staged) & 18.01 & 40.23 & 47.99 & 34.99 & 14.24 & 64.07 \\
T$\rightarrow$R (staged) & 7.03 & 17.47 & 37.29 & 18.35 & 5.15 & 56.64 \\
T+I joint (without R) & 20.58 & 39.66 & 56.92 & 41.68 & 14.92 & 62.90 \\
T+R joint (without I) & 20.58 & 40.63 & 63.13 & 48.03 & 16.81 & 62.80 \\
\textbf{Full HPD (T+I+R joint)} & \textbf{22.98} & \textbf{41.22} & \textbf{63.79} & \textbf{49.23} & \textbf{18.70} & \textbf{64.95} \\
\bottomrule
\end{tabular}
\end{table}

Using HPD-T as the anchor, the two-view rows distinguish the contributions
of I and R before combining all three views. T+R improves Step-SR from
59.25 to 63.13 and RC from 43.74 to 48.03 while TS stays at 20.58;
T+I also leaves TS unchanged and lowers both execution metrics. Only the
joint T/I/R run raises TS beyond HPD-T, suggesting that I and R contribute
to completion in combination rather than as independent additions.

Training order matters: T$\rightarrow$R drops TS from 20.58 to 7.03 and RC
from 43.74 to 18.35; T$\rightarrow$I also degrades closed-loop performance.
At matched updates and per-view exposure, both joint pairs outperform their
staged counterparts on TS and RC. Interleaving retains replay-action
supervision throughout training, while staging ends on an auxiliary target.

\subsection{Token distributions and state-conditioned action alignment}
\label{sec:token-distributions}

At the same replayed decision boundaries, we examine three aspects of action
formation: confidence in the reference control entry, uncertainty over the
reference-conditioned continuation, and validity of the action generated
without a reference prefix. Figure~\ref{fig:token-distributions} compares
Base, SFT, GRPO, and HPD; Appendix~\ref{app:token-distribution-metrics}
defines the measurements.

\begin{figure}[!htbp]
\centering
\includegraphics[width=\linewidth]{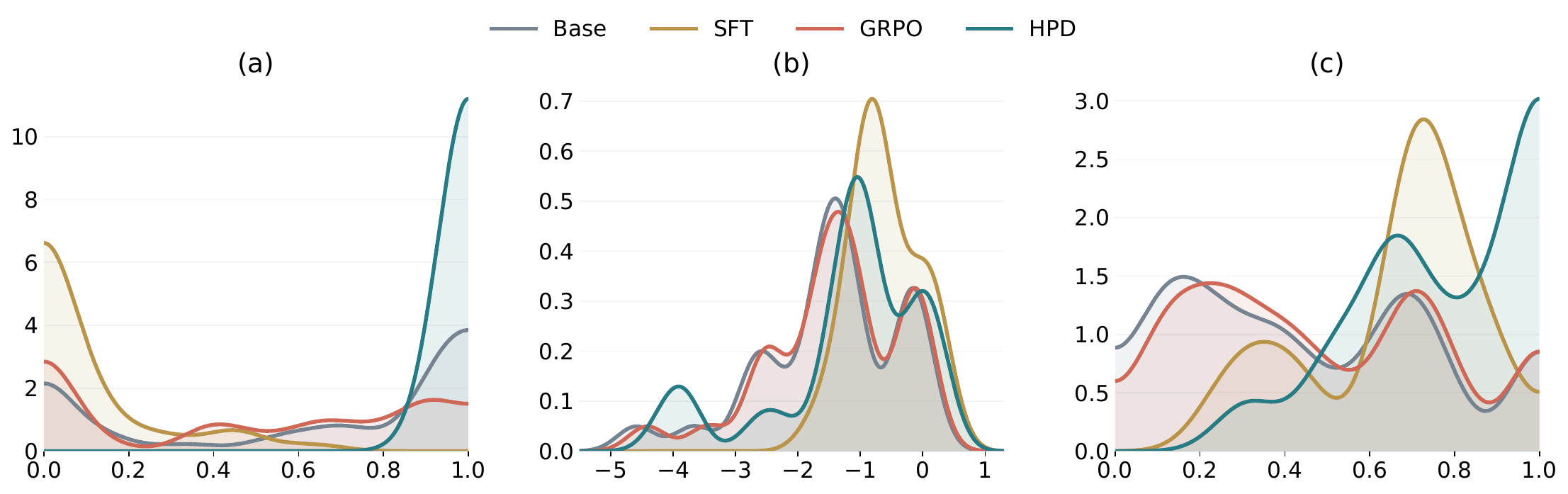}
\caption{\textbf{Control entry and reference-conditioned continuation.}
At identical harness states: (a) probability of the reference
harness control token; (b) $\log_{10}$ of mean
full-vocabulary predictive entropy in bits; (c) geometric mean
reference-token probability. Continuation statistics use reference prefixes,
exclude the initial control token, and cover at most 63 positions.
KDEs share a bandwidth across models within each panel and integrate to one
in the plotted coordinate.}
\label{fig:token-distributions}
\end{figure}

In panel (a), HPD assigns a mean probability of 0.9962 to the reference
control entry, compared with 0.6242 for Base, 0.5087 for GRPO, and 0.1296
for SFT. This is alignment with the reference entry at a supplied boundary;
SFT can score higher on rollout compliance while assigning less probability
to that particular entry.

Panel (b) examines the distribution of continuation entropy across those
boundaries. SFT's density shifts toward higher entropy and loses much of the
low-entropy mass visible for Base. HPD and GRPO retain substantial overlap
with Base, even as HPD's reference control-entry confidence rises sharply.
HPD's stronger execution alignment thus coexists with a broadly overlapping
reference-conditioned entropy profile, rather than a uniform concentration
of continuation predictions.

In panel (c), SFT and HPD assign similar median probabilities to reference
continuation tokens under the correct prefixes (0.7038 and 0.6966).
When they generate complete actions from their own prefixes, however, the
state-validity rates in this diagnostic are 7.40\% and 81.48\%, respectively.
An action couples its control entry to a continuation generated from the
model's own earlier tokens. Neither the entry score in panel (a) nor a
reference-conditioned continuation score tests that complete action. HPD's
T view fits verifier-approved actions, while I/R tie those actions to the
runtime state during training. The contrast echoes the main comparison: SFT's
Step-SR and RC improve, but F1 and TS fall. Free rollout tests whether this
state--action link holds across successive decisions.

\FloatBarrier
\Needspace{5\baselineskip}
\subsection{Adaptation across architectures and harnesses}
\label{sec:portability}

Figure~\ref{fig:cross-model-harness} shows higher TS and F1 across all evaluated
architectures and harnesses after separate adaptation. Each within-study
Base--HPD pair measures the effect of applying the same construction and
training procedure to that configuration. Llama3-8B gains 20.07 TS points
(Base TS: 0),
while Gemma4-12B \citep{gemmateam2026gemma4} gains 16.47 points and
Qwen3-30B-A3B \citep{yang2025qwen3} gains 12.69 points.
Separate adapters for ReAct, Compiler, and OpenCode gain 9.44--18.35 TS
points across different scheduling, dependency, and permission semantics.
Each adaptation rebuilds replay, control documents, and verification for its
harness while retaining the T/I/R objective and native-runtime deployment.
The two axes stress different parts of HPD: the backbone comparison changes
the policy under Pi, whereas the harness comparison changes the execution
contract under Qwen3-8B. In each setting, replay and verification construct
compatible action targets, and T/I/R connects them to native execution.
The consistent gains support portability of this construction across both
model and runtime changes.

\begin{figure}[!htbp]
\centering
\includegraphics[width=\linewidth]{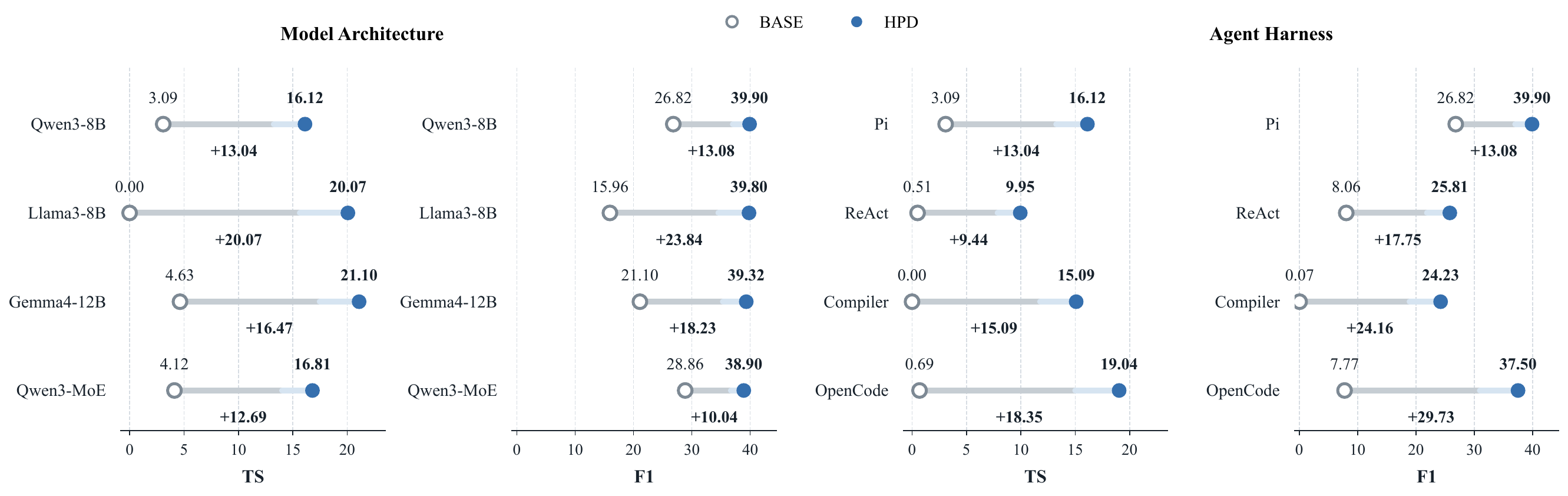}
\caption{TS and F1 across model architectures (left two panels) and separately
adapted harnesses (right two panels). All scores are percentages. The Qwen3-8B/Pi Base scores
match Table~\ref{tab:main}; its HPD scores use the 20k-optimizer-step checkpoint.}
\label{fig:cross-model-harness}
\end{figure}

\subsection{API model deployment references}
\label{sec:api-comparison}

Figure~\ref{fig:api-comparison} places HPD beside four API-accessed models
under the same Pi agent, TRAJECT-Bench tasks, and evaluation settings.
HPD reaches 22.98 TS versus 22.30 for DeepSeek V4 Flash, while GPT-5.4-mini
reaches 41.92 F1 versus 41.22 for HPD. HPD uses the same checkpoint as in
Table~\ref{tab:main}; the API rows replace only the model backend. The
different TS and F1 rankings are consistent with HPD's training target:
verifier-selected actions and control documents supervise the obligation
and termination behavior that TS measures alongside answer quality.

\begin{figure}[!htbp]
\centering
\includegraphics[width=\linewidth]{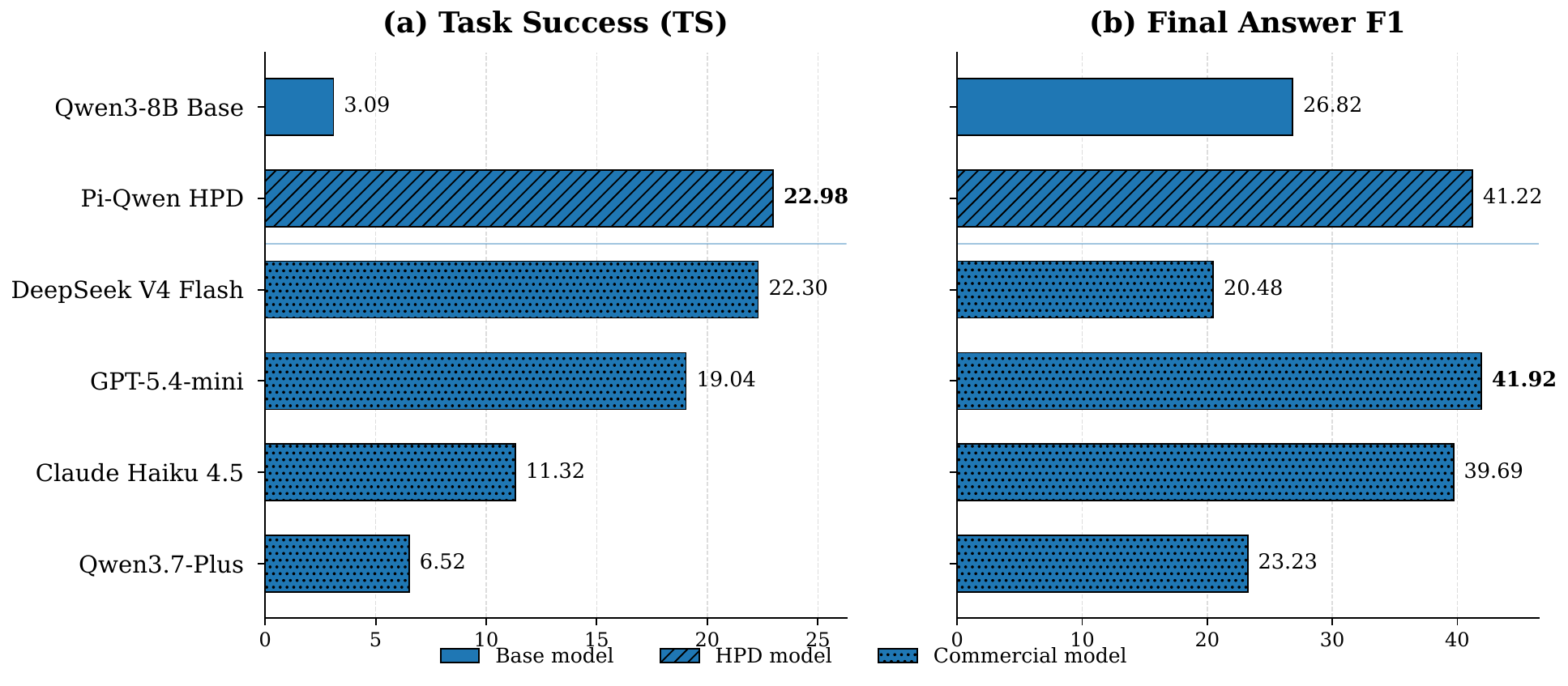}
\caption{Task success and final-answer F1 for the Qwen3-8B Base model, Pi-Qwen
HPD, and API-accessed models in the same Pi agent and evaluation setup.
Values are percentages.}
\label{fig:api-comparison}
\end{figure}

\FloatBarrier
\endgroup
\Needspace{9\baselineskip}
\section{Conclusion}
\label{sec:conclusion}

Our study addresses policy mismatch between the model and harness within an agent by targeting \emph{runtime-state mismatch}.
It teaches the model both what to do
and which execution conditions to account for. \hpd{} turns executable runtime contracts into verified action targets and explicit
control supervision and three shared-parameter views T, I, R, connect them to the native deployment policy. On TRAJECT-Bench with Pi, HPD reaches 22.98 task
success versus OPD's 15.61 at nearly equal aggregate F1. 
HPD-T already benefits from verified action supervision, while joint
control-state training further improves completion.
This benefit extends to new tasks: on 92 BFCL v4 cases without
retraining, Full HPD reaches 21.74\% task success versus HPD-T's 13.04\%.
Separate adaptations improve performance across model backbones and
harnesses. Deployment retains a single model and the existing agent
interface, without an external control document.
Together, these results support aligning agent policies not only with
task actions, but also with the runtime semantics that govern execution.

\FloatBarrier

\begingroup
\setlength{\parskip}{0pt}
\setlength{\bibsep}{2pt plus 1pt minus 0.5pt}
\bibliography{hpd_references}
\bibliographystyle{iclr2027_conference}
\endgroup
\clearpage
\appendix

\FloatBarrier
\section{Scope and Limitations}
\label{sec:limitations}

\paragraph{Harness-specific adaptation.}
\hpd{} trains policies for a specified runtime contract. Applying the method to
another harness requires reconstructing its teacher and training a separate
adapter, as in Figure~\ref{fig:cross-model-harness}. The experiments evaluate
this per-harness adaptation procedure. Reference replay supplies successful
reference prefixes as the training distribution.

\paragraph{Verifier coverage and privileged information.}
Teacher construction requires exact agreement on the designated harness-policy
and tool-call portions of the reference. The TRAJECT-Bench evaluation verifier
checks tool names, required keys, control flow, and termination using fixed
replay queues for tool outputs. Required-key matching defines its
argument-level criterion. The control document combines pre-action
observations with reference-ledger and action-witness labels. These labels
supply privileged supervision during teacher construction and training;
deployment acts directly from the native context. The component ablations
evaluate the document views as complete supervision units.

\paragraph{Online extension.}
Reference replay primarily covers successful recorded prefixes. Extending
training to states reached after deployment errors would require verified
corrections for those states.

\FloatBarrier
\section{Replay Construction Details}
\label{app:replay-details}

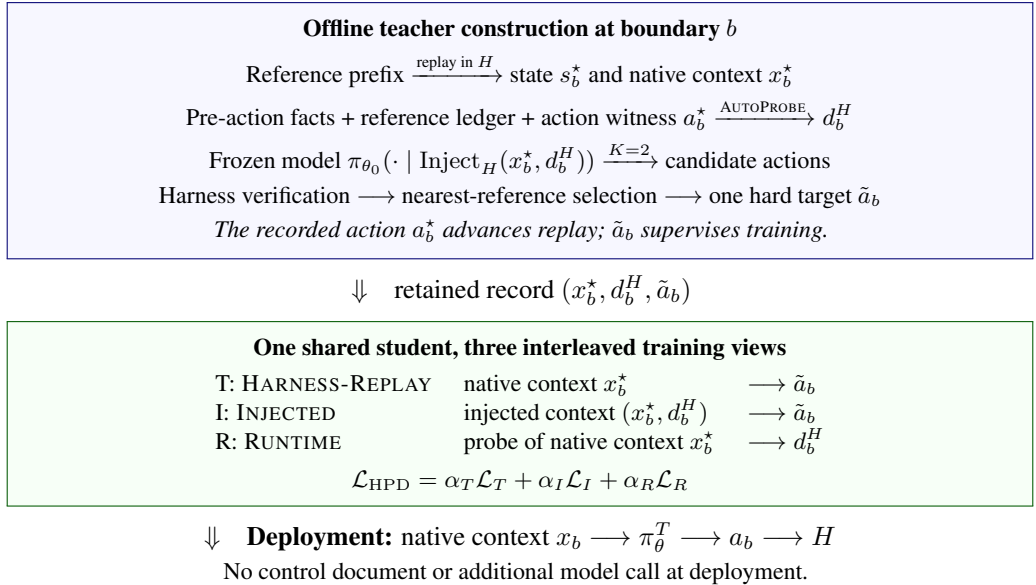
\begin{figure}[htbp]
\centering
\input{hpd_overview.tex}
\caption{Overview of Harness Policy Distillation. Reference trajectories are
replayed through the target harness; at each retained model-call boundary the
runtime state is compiled into a control document, the frozen base model is
queried with that document, and the verified candidate nearest the recorded
trajectory becomes the hard action target.
One shared student is trained on native-runtime action prediction,
document-conditioned action prediction, and document reconstruction.}
\label{fig:hpd-overview}
\end{figure}

Figure~\ref{fig:hpd-overview} summarizes the workflow from verified replay
through joint training to closed-loop agent deployment.

\subsection{Assumptions and harness adaptation}

Replay requires five conditions: each recorded action must be executable under
the target harness; the pre-action runtime prefix must be available at every
model-call boundary; observed fields must be computed only from that prefix;
privileged control annotations must declare their replay provenance; and the
successor transition must be verifiable. The recorded action is one executable
witness that reconstructs the reference path. The selected verified model
candidate supplies the action-training target. At each boundary,
\autoprobe{} compiles the control
document, the frozen base model is queried with the document-injected context,
and candidate actions are checked by the harness verifier. The main replay state
is then advanced with the recorded action, while recorded
tool observations are consumed through a per-tool queue stored in runtime state.
Unconsumed future observations remain an execution fixture and are not exposed
to the model-visible prefix. The construction stores the selected verified hard
target rather than teacher logits.

Adapting \hpd{} to a new harness requires definitions of the deployment
serializer $\Omega_H$, transition law $T_H$, decision boundaries, control-state
adapter $\Phi_H$, Inject and Probe prompt constructors, and transition and
obligation verifiers. The observation adapter may inspect only messages, prior
actions, tool results, scheduler events, permissions, and exit state available
at the current boundary. The subsequent replay compiler may additionally use
the reference stage and current action witness to label ready calls and the
continue/final condition. Pi represents lifecycle phase, completed and pending
obligations, ready calls, and the continue/final condition. Compiler additionally represents
DAG readiness, dispatch, and joins; OpenCode includes permission gates and
post-tool continuation. Harness-specific adapters translate these lifecycle
structures into control documents; the same replay, verification, and
three-view objective then apply to each adapter.

\subsection{AutoProbe boundary and field provenance}

For Pi, the probe is attached to the real provider request inside the agent
loop. When the callback fires, the current action has not yet been generated,
whereas all prior assistant actions, tool results, errors, and user steering have
already been inserted and the previous turn preparation has completed.
\autoprobe{} snapshots this state before the privileged model call. The probe
location therefore coincides with an actual model decision boundary rather than
a token-level approximation; the resulting sampled action is subsequently
checked against the harness transition and reference-stage obligation ledger.

The compiler first normalizes harness-visible objects into stable fields. It
retains message roles and text, prior tool names and arguments, tool-result
content and error status, available tool schemas, and execution modes, while
discarding call identifiers, timestamps, usage metadata, streaming events, and
runtime function objects. From this pre-action snapshot it deterministically
computes observational fields such as lifecycle position, issued calls, and
successful or failed tools. A second replay-ledger stage uses the reference
stage to label completed and pending recorded requirements. Witness-derived
fields use the current recorded action to label the ready frontier and the next
continue/final decision. Each field class consequently has a distinct
supervisory role: observed fields describe the pre-action runtime,
replay-ledger fields identify progress along the reference, and witness-derived
fields annotate the next recorded control decision. The complete document
is a privileged target. \runtime{} learns its conditional prediction from
the native context, while \injected{} receives its explicit action hints.
This provenance connects the source of each label to its use in training:
the reference supplies the supervised control labels, the injected view learns
the associated action, and the native and runtime views learn their
conditionals from deployment history.

The provenance distinction used throughout the paper is summarized as follows:
\begin{center}
\begin{tabular}{@{}lll@{}}
\toprule
Field class & Source & Prefix-causal \\
\midrule
Observed & pre-action harness snapshot & yes \\
Replay ledger & reference stage structure & no \\
Witness-derived & current recorded action & no \\
\bottomrule
\end{tabular}
\end{center}

\subsection{Teacher sampling, verification, and retries}
\label{app:teacher-construction}

Each sampling attempt produces $K=2$ complete candidate actions from the
frozen base model with the control document injected. The hard gate requires
the recorded control decision, tool names, and complete normalized tool
arguments to match; tool observations are replayed using the reference
parameters. Among candidates passing this gate, selection compares the
remaining generated fields with the reference. Writing $r(a)$ for their
vector representation, the selection is
\begin{equation}
\operatorname{SelectNearGold}(\mathcal A_b^{\mathrm{ver}};a_b^\star)
\in\arg\max_{a\in\mathcal A_b^{\mathrm{ver}}}
\operatorname{cos}\!\left(r(a),r(a_b^\star)\right).
\end{equation}
The selected verified action is the hard target shared by T and I. Thus the
executable tool content remains reference-aligned while similarity selection
operates on the other generated fields.

If neither candidate passes verification, construction stops at the failed
turn and retries from a prefix extended by one recorded trajectory turn,
including its recorded tool content. Each subsequent failure extends the
prefix by one more turn before resampling. The model generates each
continuation in the harness loop, and only a verified sampled action is
retained. The first $K=2$ attempt accepts 87.90\% of raw
trajectories, and cumulative coverage reaches 100\% by the fifth attempt.
Evaluation and deployment execute autonomously.
Table~\ref{tab:teacher-construction} summarizes these settings and diagnostics.

\begin{table}[!htbp]
\centering
\small
\caption{Teacher sampling settings and recorded construction diagnostics.}
\label{tab:teacher-construction}
\begin{tabular}{@{}ll@{}}
\toprule
Setting or diagnostic & Value \\
\midrule
Candidates per sampling attempt & $K=2$ \\
Attempts to full raw-trajectory coverage & 5 \\
Sampling temperature & 1 \\
Vocabulary filter & \texttt{distribution\_top\_k=0} (full vocabulary) \\
First-attempt acceptance rate ($K=2$) & 87.90\% \\
Cumulative coverage by attempt 5 & 100\% \\
Agreement with the reference trajectory & 90.89\% \\
\bottomrule
\end{tabular}
\end{table}

\FloatBarrier
\section{Experimental Details}
\label{app:experimental-details}

\subsection{Data construction and evaluation}

The experiments use a TRAJECT-Bench partition \citep{he2026trajectbench}
with 4,703 training tasks, 584 validation tasks, and 583 held-out test tasks,
for 5,870 tasks in total. The Pi test set contains 2,140
gold-history action boundaries. Each full task is executed through a multi-turn
agent loop with replayed tool outputs. Main end-to-end evaluation uses temperature
zero and one rollout per case. The TRAJECT-Bench results are produced by
\texttt{pi-runtime} using the replay-based metric definitions in
Appendix~\ref{app:metrics}. The Qwen3-8B Base in the checkpoint and model-size
studies uses the same end-to-end scores as Table~\ref{tab:main}; the portability study reports the Base runs
associated with its respective configurations as within-panel references.

The training pool contains 17,599 retained replay boundaries. Each boundary
defines one \replayaction{} record, one \injected{} record, and one \runtime{}
record, for 52,797 boundary--view records. The T and I records share the selected
hard target $\tilde a_b$, whereas the R record has the point target $d_b^H$.
No additional task trajectory is introduced by the extra views. Contexts follow
each backbone's native budget; construction does not impose a fixed truncation
that silently removes late turns. Teacher construction follows the sampling,
exact-match verification, similarity ranking, and retry procedure in
Appendix~\ref{app:teacher-construction}.

\subsection{Training-step convention}
\label{app:step-accounting}

Each active view receives 20k batch exposures. In a joint run, one training
round cycles through the active views, with one optimizer update per view;
for T/I/R, this means three view-specific batch updates, each on 128
sequences. The completed T-only, joint two-view, and joint three-view runs
therefore have 20k rounds and log 20k, 40k, and 60k optimizer updates,
respectively. A staged two-view run instead applies 20k consecutive updates
per view (40k in total) rather than interleaving them in joint rounds.

The optimization diagnostics retain the original optimizer-step axis: 20k,
40k, and 60k updates correspond to approximately 6.7k, 13.3k, and 20k training
rounds in the full three-view run. Separate portability and signal-density
studies retain their own fixed evaluation protocols.

\subsection{Training and comparison methods}

The principal experiments use Qwen3-8B \citep{yang2025qwen3} with LoRA
adaptation \citep{hu2022lora}. The completed HPD component runs use 20k
batch exposures per active view, as defined in Appendix~\ref{app:step-accounting}.
The shared training configuration is summarized in
Table~\ref{tab:training-config}. The portability study
trains adapters for Qwen3-8B, Llama3-8B, Gemma4-12B, and Qwen3-MoE on
Pi, and separate Qwen3-8B adapters for Pi, ReAct, Compiler, and OpenCode.

\begin{table}[!htbp]
\centering
\small
\caption{Training platform and optimization settings shared by HPD and all
baseline training runs.}
\label{tab:training-config}
\resizebox{\linewidth}{!}{%
\begin{tabular}{@{}ll@{}}
\toprule
Setting & Value \\
\midrule
Accelerators & $8\times$ NVIDIA A800-SXM4-80GB (81,920 MiB each) \\
Host resources & 64 CPU cores; 1,030,000 MB RAM \\
Runtime and precision & CUDA 12.8; BF16 (bfloat16) \\
Batch size & 128 \\
Optimizer & \texttt{torch.optim.AdamW(foreach=False)} \\
Learning rate & constant $2\times10^{-6}$ \\
Schedule & no scheduler, warmup, or decay \\
Adam parameters & $\beta=(0.9,0.999)$; $\epsilon=10^{-8}$ \\
Weight decay & 0.01 \\
Gradient clipping & maximum norm 1.0 \\
Main-run training seed & 42 \\
\bottomrule
\end{tabular}
}
\end{table}

The Qwen3-8B comparison includes the unadapted Base model; standard SFT,
which maximizes the likelihood of recorded ground-truth actions under their
reference histories; DPO
\citep{rafailov2023direct}; matched-data GiGPO and
GRPO-RLVR adaptations \citep{feng2025gigpo,shao2024deepseekmath}; GSPO
\citep{zheng2025gspo}; GDPO \citep{liu2026gdpo}; a Pi adaptation of training
with harnesses (OPHSD) \citep{zhao2026trainingharnesses}; OPSD,
using privileged-context self-distillation \citep{zhao2026selfdistilled};
classic OPD \citep{agarwal2024onpolicy}; and TurnOPD \citep{zhou2026turnopd}.
\replayaction{} is HPD's native-action view. HPD-T retains
the full HPD teacher-construction pipeline and is evaluated in
Tables~\ref{tab:main}, \ref{tab:ablation}, and~\ref{tab:seed-robustness}.
All baseline training
runs share the HPD configuration in Table~\ref{tab:training-config}; their
objectives, preference construction, rollout collection, and use of privileged
information differ. Rollout costs and total compute remain method-dependent.

\paragraph{Supervision sources.}
Table~\ref{tab:main} identifies the primary training signal for each method.
SFT uses recorded ground-truth actions, and DPO uses preference labels
\citep{rafailov2023direct}. All RL variants are grouped under \emph{Task reward}; GRPO-RLVR implements
this supervision through rule-based verification. OPHSD uses a harness-augmented model
teacher \citep{zhao2026trainingharnesses}; OPSD, OPD, and TurnOPD
use model predictions as distillation supervision. For HPD, \emph{Harness
runtime} denotes the runtime control documents and harness-verified action
targets constructed from reference replay and frozen-model candidates, as
defined in Section~\ref{sec:hpd}.

\paragraph{Distillation baselines.}
OPSD distills supervision from the base model conditioned on privileged
information; OPD and TurnOPD use a privileged 32B teacher. All three use
the same \texttt{pi-runtime} evaluation pipeline as the other baselines.

\subsection{API and cross-benchmark evaluation settings}

The API comparison uses the same Pi agent, test tasks, evaluation settings,
and metric definitions as Table~\ref{tab:main}, replacing only the model
backend with the endpoint identified in Figure~\ref{fig:api-comparison}.
The HPD checkpoint is unchanged. BFCL evaluation instead replaces the test
tasks and retains each model's TRAJECT-Bench checkpoint without retraining,
as well as the same Pi agent and model-side evaluation settings. Task and
tool execution follow BFCL v4's default settings. Appendix~\ref{app:bfcl-ood}
reports this cross-benchmark transfer.

\Needspace{5\baselineskip}
\subsection{Seed-study results}
\label{app:statistical-reporting}

Table~\ref{tab:seed-robustness} reports five additional runs with seeds
715--719. These runs are separate from the main-table run
with seed 42. Cases interrupted by device crashes are excluded from the affected run's
evaluation, so effective task counts can differ across runs. Each run is
scored on its retained cases, and the table reports the mean and standard
deviation of the five run-level scores. Base is evaluated under the same
five seed settings; its variation reflects differences in inference outcomes
across those evaluations, without model training.
Full HPD reaches 22.14 mean TS, improving by 19.46 points over Base. Relative
to HPD with the T view alone, joint T/I/R training adds 3.94 points in TS,
1.29 in F1, 4.37 in Step-SR, and 5.44 in RC. Both HPD variants use the same
teacher-construction procedure; this comparison measures the additional
benefit of jointly supervising control documents and actions.

\begin{table}[!htbp]
\centering
\small
\setlength{\tabcolsep}{4pt}
\caption{Supplementary seed-study results from \texttt{pi-runtime}.
Values are percentages, reported as mean $\pm$ standard deviation over
five seeds (715--719), excluding the main-run seed 42. Run-level scores
exclude cases interrupted by device crashes. HPD-T retains
the full teacher-construction pipeline and trains only the native-action view.}
\label{tab:seed-robustness}
\begin{tabular}{@{}lrrrr@{}}
\toprule
Method & TS & F1 & Step-SR & RC \\
\midrule
Base & $2.68 \pm 0.23$ & $26.81 \pm 0.11$ & $26.54 \pm 0.13$ & $19.62 \pm 0.15$ \\
HPD-T & $18.20 \pm 1.67$ & $39.50 \pm 0.66$ & $56.21 \pm 3.43$ & $42.82 \pm 0.81$ \\
Full HPD & $\mathbf{22.14 \pm 1.18}$ & $\mathbf{40.79 \pm 0.39}$ & $\mathbf{60.58 \pm 3.54}$ & $\mathbf{48.26 \pm 0.88}$ \\
\bottomrule
\end{tabular}
\end{table}

Full HPD improves all four mean metrics over its T-only ablation. Its
TS and F1 standard deviations are 1.18 and 0.39, respectively, compared
with 1.67 and 0.66 for T only. The five-run results reinforce the component
comparison: harness-verified target construction supports native-action
learning, and joint control-document supervision improves completion,
reference-prefix execution, and compliance further.

For paired significance testing, we use a two-sided exact sign-flip test on
source-cluster differences across the five seed runs. We apply Holm correction
within each comparison to TS, F1, Step-SR, and RC. The adjusted TS $p$-values
for Full HPD versus HPD-T and OPD are 0.00343 and 0.00338, respectively.

\FloatBarrier
\subsection{Cross-benchmark transfer without retraining}
\label{app:bfcl-ood}

\begin{figure}[!htbp]
\centering
\input{hpd_bfcl_ood.tex}
\caption{BFCL v4 transfer on 92 cases without retraining. Panels show
(a) TS, (b) final-answer F1, and (c) Step-SR for the eight methods.
All axes share the same zero-based percentage scale. HPD-T and Full HPD are
blue; Full HPD is hatched. Numerical labels give the scores on this subset.}
\label{fig:bfcl-ood}
\end{figure}
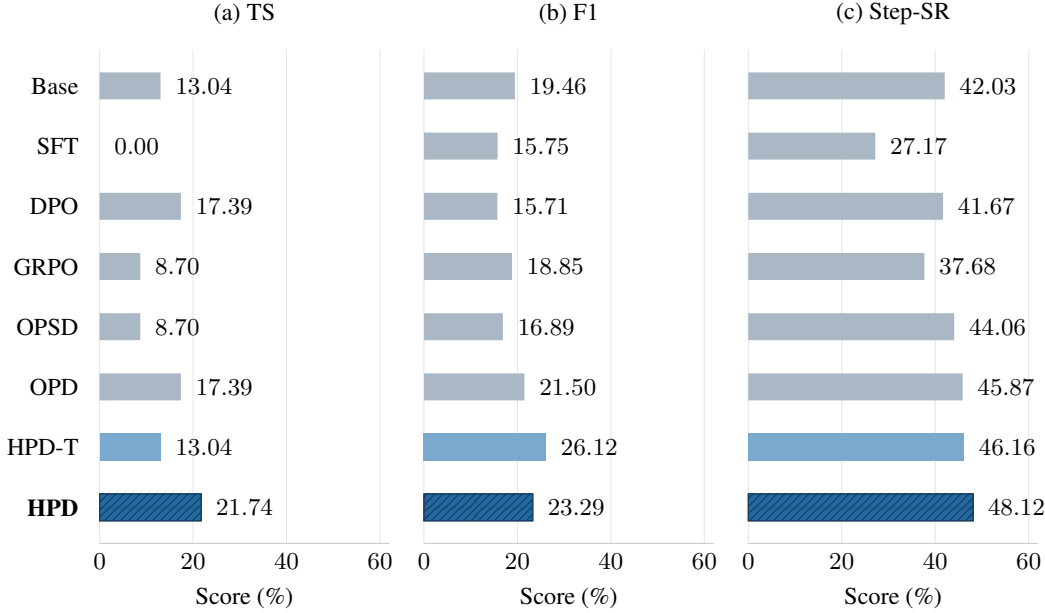

We evaluate Base, SFT, DPO, GRPO, OPD, OPSD, HPD-T, and HPD on 92 BFCL v4 cases using
the same checkpoints as in the TRAJECT-Bench comparison, without further
training on BFCL. The Pi agent and model-side evaluation settings remain
unchanged; only the test tasks are replaced. Task and tool execution follow
BFCL v4's default settings. BFCL evaluates function calling and stateful
agentic behavior \citep{patil2025bfcl}. Figure~\ref{fig:bfcl-ood} reports TS,
final-answer F1, and Step-SR, using the same metric definitions as the main
comparison. This BFCL evaluation reports these three metrics.

Full HPD has the highest TS and Step-SR (21.74 and 48.12), whereas HPD-T has
the highest F1 (26.12). HPD-T matches Base's 13.04 TS but exceeds its F1 and
Step-SR by 6.66 and 4.13 points. Relative to HPD-T, Full HPD gains 8.70 TS
and 1.96 Step-SR points while losing 2.83 F1 points. T-only replay thus
transfers answer overlap and prefix execution without improving completed-task
coverage on this collection; the additional I/R views are associated with
the completion gain. OPD and DPO tie for the highest external-baseline TS at
17.39. OPD has the strongest external-baseline F1 and Step-SR (21.50 and
45.87); Full HPD exceeds it by 4.35, 1.79, and 2.25 points, respectively.
OPSD reaches 8.70 TS, 16.89 F1, and 44.06 Step-SR, improving reference-prefix
execution over Base without improving task completion. DPO improves TS over
Base but lowers F1 and Step-SR; SFT and GRPO fall below Base on all three
metrics. These differences suggest that verified replay alone can transfer
partial execution behavior, while I/R better connects those actions to the
runtime control state required for completion on new tasks.

\FloatBarrier
\section{Optimization Diagnostics}

\subsection{Checkpoint dynamics}
\label{app:checkpoints}

\begin{figure}[!htbp]
\centering
\includegraphics[width=\linewidth]{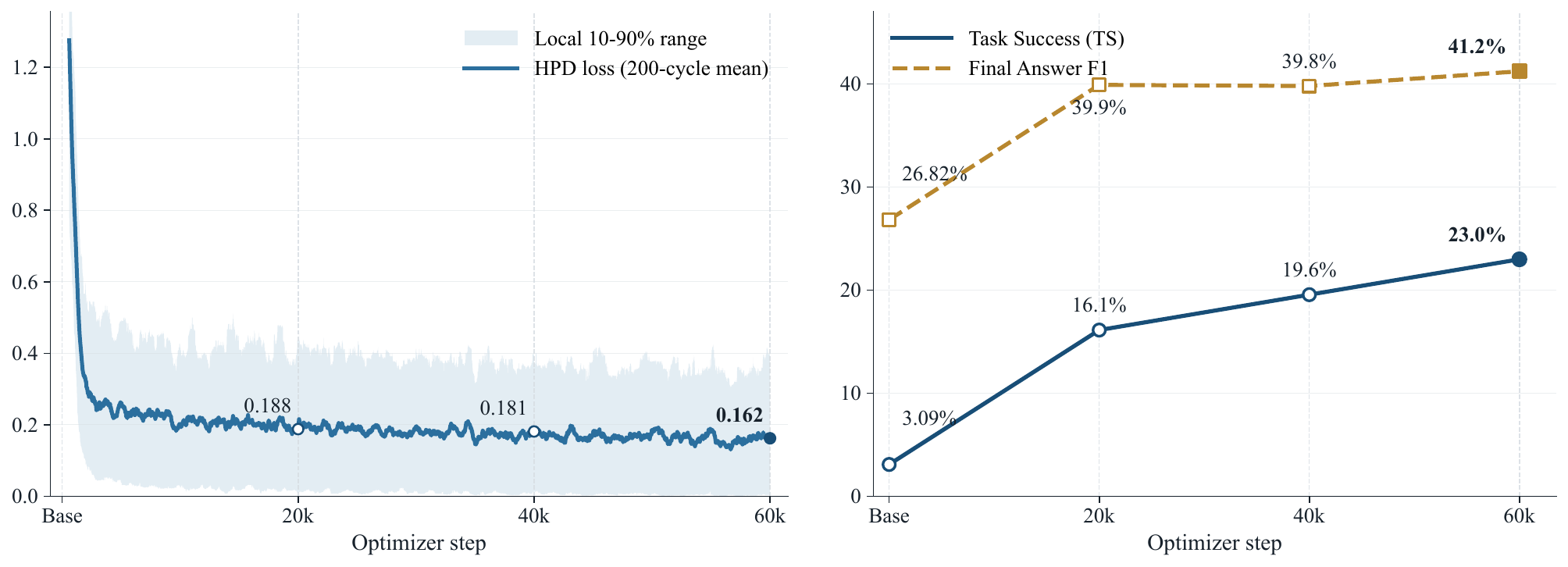}
\caption{Full HPD optimization over 60k optimizer steps (20k training rounds).
Left: 200-cycle mean loss and local 10--90\% range. Right: task success and
final-answer F1. Base scores are TS 3.09\% and F1 26.82\%;
trained-checkpoint labels are rounded to one decimal.}
\label{fig:training-curves}
\end{figure}

Figure~\ref{fig:training-curves} aligns optimization loss with end-to-end
performance at 20k, 40k, and 60k optimizer steps in the full three-view run.
The 200-cycle mean loss decreases from 0.188 to
0.181 and 0.162. TS rises from 3.09 to 16.12, 19.55, and 22.98, while F1 rises from 26.82
to 39.90, 39.79, and 41.22. Continued training improves executable
cross-turn completion after most answer-content gains have appeared.

\begin{table}[!htbp]
\centering
\small
\caption{Intermediate and completed full-HPD checkpoints evaluated by
\texttt{pi-runtime}, indexed by optimizer steps. The 60k endpoint completes 20k training
rounds. Scores are percentages.}
\label{tab:checkpoint-details}
\resizebox{\linewidth}{!}{%
\begin{tabular}{@{}lrrrrrr@{}}
\toprule
Optimizer steps & TS & F1 & Action-Acc & Step-SR & RC & Exact \\
\midrule
40k (intermediate) & 19.55 & 39.79 & 63.74 & 58.27 & 43.40 & 15.61 \\
60k (completed) & 22.98 & 41.22 & 64.95 & 63.79 & 49.23 & 18.70 \\
\bottomrule
\end{tabular}}
\end{table}

\FloatBarrier
\subsection{Loss decomposition}

Because each optimizer update uses one view, a naive moving average mixes
objective composition with optimization progress. Aggregating each complete
T/I/R cycle yields the trend in Figure~\ref{fig:training-curves}. At the 60k-optimizer-step endpoint, the
view losses are 0.2415 for \replayaction{}, 0.2283 for \injected{}, and
0.0170 for \runtime{}. The losses summarize different prediction tasks with completion-length
normalization. In particular, the document loss measures token prediction
across observed, ledger, and witness-derived fields. Reporting the losses
separately tracks optimization progress for each supervision path;
rollout metrics measure the resulting policy's execution behavior.

\subsection{Signal-density search}

Figure~\ref{fig:signal-density} and Table~\ref{tab:signal-density} report a
separate Qwen3-8B diagnostic at 20k optimizer steps per configuration.
Harness-Replay is fixed at 1.0 while either Injected or Runtime is varied over
0.2, 0.6, 1.4, and 1.8; the other auxiliary view stays at 1.0. These weights
set the sampling frequency of each view. At the fixed update budget, changing
one frequency reallocates updates among the three supervision paths. The
default 1:1:1 cycle anchors both sweeps.
Injected at 0.2 gives the highest RC (43.57) and Step-SR (56.40), gains of
2.92 and 3.64 percentage points over the default. Runtime at 0.6 gives the
highest TS (18.18), Exact (14.24), and F1 (40.05), improving by 2.06, 1.03,
and 0.15 points, respectively. The preferred frequency depends on the target
metric: Injected at 0.2 favors prefix execution and compliance, whereas
Runtime at 0.6 favors completed tasks in this sweep. Each axis is searched
separately with one seed, giving a coordinate-wise comparison around the
default cycle.

\begingroup
\setlength{\intextsep}{10pt}
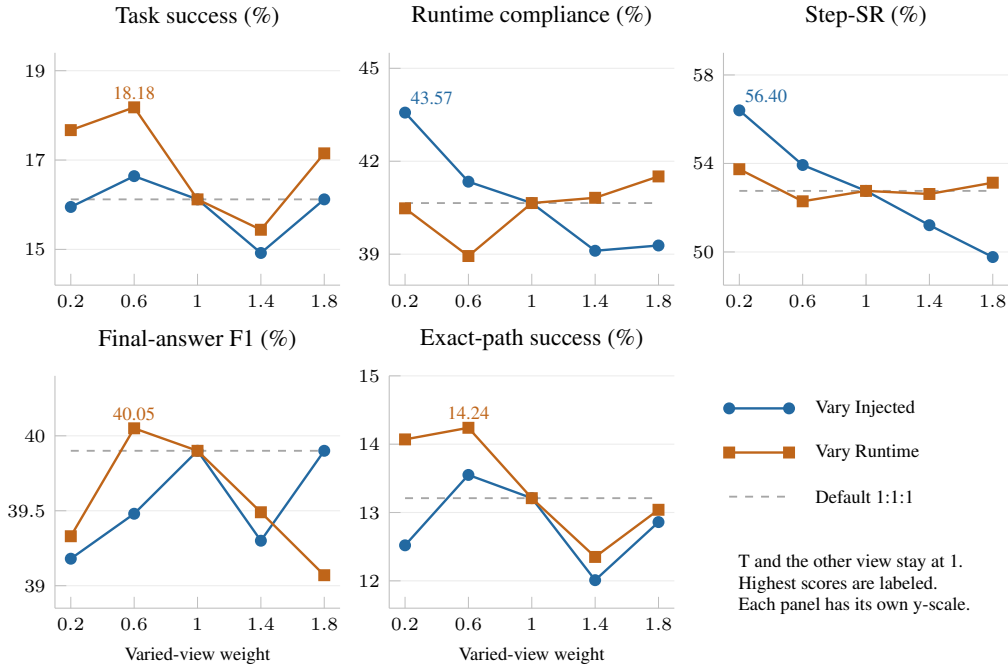
\begin{figure}[!htbp]
\centering
\begingroup
\definecolor{hpdInjected}{HTML}{2369A1}
\definecolor{hpdRuntime}{HTML}{BF6519}
\pgfplotstableread{
weight its rts irc rrc istep rstep ifone rfone iexact rexact
0.2 15.95 17.67 43.57 40.48 56.40 53.74 39.18 39.33 12.52 14.07
0.6 16.64 18.18 41.34 38.94 53.93 52.29 39.48 40.05 13.55 14.24
1.0 16.12 16.12 40.65 40.65 52.76 52.76 39.90 39.90 13.21 13.21
1.4 14.92 15.44 39.11 40.82 51.21 52.62 39.30 39.49 12.01 12.35
1.8 16.12 17.15 39.28 41.51 49.77 53.13 39.90 39.07 12.86 13.04
}\hpdsweepdata
\pgfplotsset{
  hpd injected/.style={color=hpdInjected,mark=*,mark size=1.8pt,line width=0.9pt},
  hpd runtime/.style={color=hpdRuntime,mark=square*,mark size=1.8pt,line width=0.9pt},
  hpd default/.style={color=gray!70,dashed,line width=0.7pt,no marks},
  /tikz/hpd peak/.style={font=\scriptsize,anchor=south,yshift=2pt,fill=white,inner sep=1pt},
}
\begin{tikzpicture}
\begin{groupplot}[
  group style={group size=3 by 2,horizontal sep=0.65cm,vertical sep=1.2cm},
  scale only axis,width=0.27\linewidth,height=0.22\linewidth,
  xmin=0.1,xmax=1.9,xtick={0.2,0.6,1.0,1.4,1.8},
  xticklabel style={font=\scriptsize},yticklabel style={font=\scriptsize},
  title style={font=\small},label style={font=\scriptsize},
  xlabel={Varied-view weight},axis lines*=left,
  axis line style={gray!50},tick style={gray!50},
  ymajorgrids,grid style={gray!15},clip=false,
]
\nextgroupplot[title={Task success (\%)},xlabel={},ymin=14.2,ymax=19.4,ytick={15,17,19}]
\addplot[hpd default] coordinates {(0.2,16.12) (1.8,16.12)};
\addplot[hpd injected] table[x=weight,y=its] {\hpdsweepdata};
\addplot[hpd runtime] table[x=weight,y=rts] {\hpdsweepdata};
\node[hpd peak,text=hpdRuntime] at (axis cs:0.6,18.18) {18.18};
\nextgroupplot[title={Runtime compliance (\%)},xlabel={},ymin=38,ymax=45.5,ytick={39,42,45}]
\addplot[hpd default] coordinates {(0.2,40.65) (1.8,40.65)};
\addplot[hpd injected] table[x=weight,y=irc] {\hpdsweepdata};
\addplot[hpd runtime] table[x=weight,y=rrc] {\hpdsweepdata};
\node[hpd peak,text=hpdInjected,anchor=south west,xshift=1pt] at (axis cs:0.2,43.57) {43.57};
\nextgroupplot[title={Step-SR (\%)},xlabel={},ymin=48.5,ymax=59,ytick={50,54,58}]
\addplot[hpd default] coordinates {(0.2,52.76) (1.8,52.76)};
\addplot[hpd injected] table[x=weight,y=istep] {\hpdsweepdata};
\addplot[hpd runtime] table[x=weight,y=rstep] {\hpdsweepdata};
\node[hpd peak,text=hpdInjected,anchor=south west,xshift=1pt] at (axis cs:0.2,56.40) {56.40};
\nextgroupplot[title={Final-answer F1 (\%)},ymin=38.85,ymax=40.4,ytick={39,39.5,40}]
\addplot[hpd default] coordinates {(0.2,39.90) (1.8,39.90)};
\addplot[hpd injected] table[x=weight,y=ifone] {\hpdsweepdata};
\addplot[hpd runtime] table[x=weight,y=rfone] {\hpdsweepdata};
\node[hpd peak,text=hpdRuntime] at (axis cs:0.6,40.05) {40.05};
\nextgroupplot[title={Exact-path success (\%)},ymin=11.6,ymax=15,ytick={12,13,14,15}]
\addplot[hpd default] coordinates {(0.2,13.21) (1.8,13.21)};
\addplot[hpd injected] table[x=weight,y=iexact] {\hpdsweepdata};
\addplot[hpd runtime] table[x=weight,y=rexact] {\hpdsweepdata};
\node[hpd peak,text=hpdRuntime] at (axis cs:0.6,14.24) {14.24};
\nextgroupplot[hide axis,xmin=0,xmax=1,ymin=0,ymax=1,xlabel={}]
\addplot[hpd injected] coordinates {(0.02,0.86) (0.23,0.86)};
\node[anchor=west,font=\scriptsize] at (axis cs:0.29,0.86) {Vary Injected};
\addplot[hpd runtime] coordinates {(0.02,0.67) (0.23,0.67)};
\node[anchor=west,font=\scriptsize] at (axis cs:0.29,0.67) {Vary Runtime};
\addplot[hpd default] coordinates {(0.02,0.48) (0.23,0.48)};
\node[anchor=west,font=\scriptsize] at (axis cs:0.29,0.48) {Default 1:1:1};
\node[anchor=north west,font=\scriptsize,align=left] at (axis cs:0.02,0.27)
  {T and the other view stay at 1.\\ Highest scores are labeled.\\ Each panel has its own y-scale.};
\end{groupplot}
\end{tikzpicture}
\endgroup
\caption{Single-axis sampling-weight search at 20k optimizer steps. Blue circles
vary Injected; orange squares vary Runtime. Dashed lines mark the shared
1:1:1 default. Each panel shows one metric with its own y-scale and labels its
maximum. Lines connect evaluated settings.}
\label{fig:signal-density}
\end{figure}

\begin{table}[!htbp]
\centering
\small
\caption{Complete single-axis sampling-weight search. Ratios give the realized
T:I:R sampling cycle; all scores are percentages. Bold marks the highest score
in each metric.}
\label{tab:signal-density}
\begin{tabular}{@{}llrrrrr@{}}
\toprule
Varied view & T:I:R & TS & F1 & Step-SR & RC & Exact \\
\midrule
Default & 1:1:1 & 16.12 & 39.90 & 52.76 & 40.65 & 13.21 \\
\midrule
Injected (0.2) & 5:1:5 & 15.95 & 39.18 & \textbf{56.40} & \textbf{43.57} & 12.52 \\
Injected (0.6) & 5:3:5 & 16.64 & 39.48 & 53.93 & 41.34 & 13.55 \\
Injected (1.4) & 5:7:5 & 14.92 & 39.30 & 51.21 & 39.11 & 12.01 \\
Injected (1.8) & 5:9:5 & 16.12 & 39.90 & 49.77 & 39.28 & 12.86 \\
\midrule
Runtime (0.2) & 5:5:1 & 17.67 & 39.33 & 53.74 & 40.48 & 14.07 \\
Runtime (0.6) & 5:5:3 & \textbf{18.18} & \textbf{40.05} & 52.29 & 38.94 & \textbf{14.24} \\
Runtime (1.4) & 5:5:7 & 15.44 & 39.49 & 52.62 & 40.82 & 12.35 \\
Runtime (1.8) & 5:5:9 & 17.15 & 39.07 & 53.13 & 41.51 & 13.04 \\
\bottomrule
\end{tabular}
\end{table}
\endgroup
\FloatBarrier
\Needspace{5\baselineskip}
\subsection{Model-size scaling study}

Table~\ref{tab:model-size} evaluates Base and HPD across seven Qwen3
checkpoints. HPD improves both TS and F1 at every size, with TS gains
ranging from 4.46 points at 0.6B to 14.41 at 14B; the 32B gain is 13.90
points. Step-SR and RC also improve at every size. The 14B and 32B dense
checkpoints exceed the 30B-A3B MoE checkpoint on TS. All rows use the same
583 cases and one seed.

Each Base evaluation contains 583 unique cases and 2,140 reference steps,
with no worker failures. TS uses the historical task-success definition;
F1, RC, and Exact use the runtime evaluation summaries. Action-Acc comes
from a separate native gold-history boundary evaluation. The Qwen3-8B HPD row corresponds to the 20k-optimizer-step
checkpoint, rather than the completed 60k checkpoint in Table~\ref{tab:main}.

\begin{table}[!htbp]
\centering
\small
\caption{Qwen3 model-size results. Action-Acc uses the native gold-history
boundary protocol separately from end-to-end evaluation. Exact retains the historical
strict trajectory-success definition, not answer-string exact match.
All values are percentages.}
\label{tab:model-size}
\resizebox{\linewidth}{!}{%
\begin{tabular}{@{}llrrrrrr@{}}
\toprule
Model & Variant & TS & F1 & Action-Acc & Step-SR & RC & Exact \\
\midrule
Qwen3-0.6B & Base & 0.00 & 24.78 & 52.06 & 3.64 & 0.00 & 0.00 \\
 & HPD & 4.46 & 25.70 & 54.07 & 24.20 & 16.50 & 2.40 \\
Qwen3-1.7B & Base & 0.00 & 27.00 & 49.90 & 4.53 & 0.00 & 0.00 \\
 & HPD & 6.52 & 30.50 & 59.00 & 33.60 & 20.80 & 3.30 \\
Qwen3-4B & Base & 1.03 & 20.81 & 53.40 & 9.81 & 7.38 & 0.86 \\
 & HPD & 15.09 & 38.50 & 59.40 & 48.60 & 38.90 & 11.50 \\
Qwen3-8B & Base & 3.09 & 26.82 & 53.40 & 26.45 & 19.90 & 2.57 \\
 & HPD & 16.12 & 39.90 & 60.80 & 52.80 & 40.70 & 13.21 \\
Qwen3-14B & Base & 3.43 & 27.96 & 52.90 & 14.77 & 15.61 & 2.74 \\
 & HPD & 17.84 & 40.00 & 60.00 & 57.40 & 44.90 & 14.92 \\
Qwen3-30B-A3B & Base & 4.12 & 28.86 & 52.90 & 14.44 & 17.67 & 2.74 \\
 & HPD & 16.81 & 38.90 & 58.80 & 53.60 & 40.10 & 13.21 \\
Qwen3-32B & Base & 6.00 & 30.42 & 56.82 & 19.63 & 25.73 & 4.12 \\
 & HPD & 19.90 & 40.29 & 62.15 & 64.49 & 49.23 & 16.81 \\
\bottomrule
\end{tabular}%
}
\end{table}
\FloatBarrier
\section{Metric Definitions}
\label{app:metrics}

Unless noted otherwise, reported values are percentages. Final-answer F1 is
token-level F1 between the emitted final response and the reference answer.

\paragraph{Gold-history action accuracy.}
At each of the 2,140 Pi provider boundaries, the evaluator supplies the
recorded gold history as context and compares the sequence likelihood of the
reference next action with that of a constructed alternative. Action-Acc is the
fraction of boundaries where the reference action ranks higher. The two
candidates are scored but not executed; each boundary is evaluated
independently, so model-generated rollout histories do not enter this metric.
The rollout metrics below evaluate the policy through execution.

\paragraph{Reference-prefix step success.}
Step-SR counts correct reference steps until first divergence:
\begin{equation}
\mathrm{Step\mbox{-}SR}=
\frac{\sum_i \text{correct prefix steps in rollout }i}
{\sum_i \text{reference steps in task }i}.
\end{equation}
Later actions receive no reference-path credit after the first divergence
by definition, even if the rollout later recovers or reaches an equivalent
runtime state.

\paragraph{Runtime compliance, Exact, and task success.}
RC is the fraction of tasks satisfying the harness execution contract: required
tool sequence and parameter keys, exactly one final response, native exit, and
a healthy worker. Exact additionally requires reproduction of the reference
control path and final-answer F1 of at least 0.5. For task $i$, let $O_i$
indicate eventual obligation completion, $F_i$ final-answer F1, $C_i$ exactly
one final, $E_i$ native exit, and $W_i$ healthy worker execution:
\begin{equation}
\mathrm{TS}_i=\mathbf 1[O_i]\mathbf 1[F_i\geq0.5]
\mathbf 1[C_i]\mathbf 1[E_i]\mathbf 1[W_i].
\end{equation}
TS is the mean of this indicator. A recovered success can have Exact$=0$
because retries or a valid action ordering different from the reference path
are allowed. The TRAJECT-Bench evaluation contract uses required parameter keys as its
argument-level criterion and replayed observations as tool outputs. RC, Exact, and TS
therefore describe execution within this specified runtime.

\subsection{Token-distribution diagnostic}
\label{app:token-distribution-metrics}
At boundary $b$, let $c_b^\star$ be the correct initial control token and
$u_{b,1:m_b}^\star$ the reference continuation after it, truncated to
$m_b\leq63$ tokens. For $m_b>0$, define
\begin{equation}
\begin{aligned}
p_{b,j}(v)
&=\pi_\theta(v\mid x_b^\star,c_b^\star,u_{b,<j}^\star),\\
\bar h_b
&=-\frac{1}{m_b}\sum_{j=1}^{m_b}\sum_{v\in\mathcal V}
p_{b,j}(v)\log_2 p_{b,j}(v),\\
g_b
&=\exp\!\left(\frac{1}{m_b}\sum_{j=1}^{m_b}
\log p_{b,j}(u_{b,j}^\star)\right),
\end{aligned}
\label{eq:continuation-diagnostics}
\end{equation}
where $\mathcal V$ is the full token vocabulary. The three panels of
Figure~\ref{fig:token-distributions} plot
$\pi_\theta(c_b^\star\mid x_b^\star)$, $\log_{10}\bar h_b$, and $g_b$,
respectively. Thus panel (b) takes the logarithm \emph{after} averaging
position-wise entropy, while panel (c) exponentiates the mean reference-token
log probability. Neither continuation statistic uses model-generated prefixes.

\subsection{Interpreting F1 and task completion}
\label{app:f1-interpretation}

For task-averaged final-answer F1, let $S_i$ be the binary task-success
indicator and $F_i\in[0,1]$ the final-answer F1 for task $i$. Writing
$p_{\mathrm{TS}}=\mathbb E[S_i]$, the mean decomposes as
\begin{equation}
\mathbb E[F_i]
=p_{\mathrm{TS}}\,\mathbb E[F_i\mid S_i=1]
+(1-p_{\mathrm{TS}})\,\mathbb E[F_i\mid S_i=0].
\label{eq:f1-completion-decomposition}
\end{equation}
Thus aggregate F1 can change through the fraction of successful tasks,
the mean F1 within either group, or both. The tables provide the aggregate
and success rate, which describe answer overlap and completed-task coverage
as separate outcomes.

HPD and OPD have nearly equal aggregate F1 (41.22 versus 41.04), while HPD
has substantially higher TS (22.98 versus 15.61) and Exact (18.70 versus
14.07). OPSD similarly raises TS over Base (13.55 versus 3.09)
without raising F1 (26.27 versus 26.82). Together with HPD's T-only and
joint-view ablations, these patterns identify HPD's completion advantage at
comparable aggregate answer overlap. The distinction follows directly from
the metrics: F1 averages answer overlap over all tasks, whereas TS counts
tasks that jointly satisfy the execution conditions and answer threshold.
Similar mean F1 can therefore accompany different successful-completion
coverage.

Equation~\ref{eq:f1-completion-decomposition} also specifies a task-level
analysis for attributing F1 changes: measure answer overlap on tasks
completed by both methods, newly completed tasks, previously completed tasks
that become unsuccessful, and tasks completed by neither method.
Grouping instead by execution-only conditions, such as obligation completion
and native termination, separates execution coverage from the answer threshold
used in TS. These two decompositions distinguish the coverage of successful
execution from answer quality within each group. Runtime alignment and
reference-distribution fitting can contribute jointly; paired,
execution-conditioned analyses separate their completion-coverage and
within-group answer-quality outcomes.

\FloatBarrier
\section{Distillation Derivation and Implementation}
\label{app:derivation}

\subsection{Replay-teacher distributions}
Offline construction fixes a document and selected action at each retained
boundary, defining the point-mass teacher. Here $\delta_z$ assigns probability
one to the sequence $z$ and zero elsewhere; $d$ and $a$ are document and action
variables, respectively:
\begin{equation}
\begin{aligned}
\pi_b^{\mathrm{rep}}(d,a\mid x_b^\star)
&=\delta_{d_b^H}(d)\;\delta_{\tilde a_b}(a).
\end{aligned}
\label{eq:joint-replay-teacher}
\end{equation}
Write $\pi^{\mathrm{rep}}(d,a\mid x)$ for the empirical conditional
distribution of document--action pairs given context $x$. Its document
marginal $\pi_R^{\mathrm{rep}}$, document-conditioned action distribution
$\pi_I^{\mathrm{rep}}$, and native action marginal $\pi_T^{\mathrm{rep}}$ satisfy
\begin{align}
\pi^{\mathrm{rep}}(d,a\mid x)
&=\pi_R^{\mathrm{rep}}(d\mid x)
  \pi_I^{\mathrm{rep}}(a\mid x,d),\label{eq:teacher-factorization}\\
\pi_T^{\mathrm{rep}}(a\mid x)
&=\sum_d\pi^{\mathrm{rep}}(d,a\mid x).
\label{eq:teacher-marginal}
\end{align}
The document includes runtime facts and privileged replay labels. Each record
has a point target; records sharing a serialized context induce an empirical
mixture when conditioned on that context. This distinction connects stored
supervision to the conditional teacher objectives.

\subsection{Conditional KL decomposition}

All KL divergences use natural logarithms. Expectations below are taken
under the empirical teacher conditionals defined above.

\runtime{} and \injected{} define the factorized student
$\widetilde\pi_\theta(d,a\mid x)=
\pi_\theta^R(d\mid x)\pi_\theta^I(a\mid x,d)$. Suppressing boundary and
harness indices, the KL chain rule gives
\begin{equation}
\begin{aligned}
&D_{\mathrm{KL}}\!\left(
\pi^{\mathrm{rep}}(d,a\mid x)\,\Vert\,
\widetilde\pi_\theta(d,a\mid x)\right)\\
&\quad=D_{\mathrm{KL}}\!\left(
\pi_R^{\mathrm{rep}}(\cdot\mid x)\,\Vert\,
\pi_\theta^R(\cdot\mid x)\right)\\
&\qquad+\mathbb E_{d\sim\pi_R^{\mathrm{rep}}(\cdot\mid x)}
D_{\mathrm{KL}}\!\left(
\pi_I^{\mathrm{rep}}(\cdot\mid x,d)\,\Vert\,
\pi_\theta^I(\cdot\mid x,d)\right).
\end{aligned}
\label{eq:kl-chain}
\end{equation}
The chain rule motivates the R and I losses; T additionally matches the
teacher's native-runtime action marginal. The three losses in
Equation~\ref{eq:hpd-objective} are
\begin{equation}
\begin{aligned}
\mathcal L_R
&=\mathbb E_{x\sim\mu_H^{\mathrm{rep}}}
D_{\mathrm{KL}}\!\left(
\pi_R^{\mathrm{rep}}(\cdot\mid x)\,\Vert\,
\pi_\theta^R(\cdot\mid x)\right),\\
\mathcal L_I
&=\mathbb E_{x\sim\mu_H^{\mathrm{rep}}}
\mathbb E_{d\sim\pi_R^{\mathrm{rep}}(\cdot\mid x)}
D_{\mathrm{KL}}\!\left(
\pi_I^{\mathrm{rep}}(\cdot\mid x,d)\,\Vert\,
\pi_\theta^I(\cdot\mid x,d)\right),\\
\mathcal L_T
&=\mathbb E_{x\sim\mu_H^{\mathrm{rep}}}
D_{\mathrm{KL}}\!\left(
\pi_T^{\mathrm{rep}}(\cdot\mid x)\,\Vert\,
\pi_\theta^T(\cdot\mid x)\right),
\end{aligned}
\label{eq:conditional-view-losses}
\end{equation}
where $\pi_T^{\mathrm{rep}}$ is the deployment marginal of
Equation~\ref{eq:teacher-marginal}.

\subsection{Stored targets and the implemented loss}

After sampling, verification, and near-gold selection, let $c_b^v$ denote the
stored prompt and $p_b^v$ the target-sequence distribution for boundary $b$
and view $v\in\{T,I,R\}$. The three implementation pairs are
\begin{equation}
\begin{aligned}
(c_b^T,p_b^T)
&=\left(x_b^\star,\delta_{\tilde a_b}\right),\\
(c_b^I,p_b^I)
&=\left(\operatorname{Inject}_H(x_b^\star,d_b^H),
\delta_{\tilde a_b}\right),\\
(c_b^R,p_b^R)
&=\left(\operatorname{Probe}_H(x_b^\star),\delta_{d_b^H}\right).
\end{aligned}
\label{eq:view-pairs}
\end{equation}
Thus the two action views share the selected verifier-approved hard target,
while the document view predicts the control document stored for the same
boundary. Student training consumes these stored targets and requires no
teacher forward pass. Rejected and unselected candidates do not enter the
three supervised targets.

\paragraph{Per-record NLL and conditional teacher KL.}
Equation~\ref{eq:hard-target-kl} establishes the point-target identity for
each complete stored sequence. To connect these per-record losses to the
conditional teacher, average over the empirical distribution of prompt--target
pairs.

For any frozen target distribution $p$ and context $c$, with $\mathcal H(p)$
denoting Shannon entropy using natural logarithms,
\begin{equation}
\mathbb E_{y\sim p}\left[-\log\pi_\theta(y\mid c)\right]
=D_{\mathrm{KL}}\!\left(p\,\Vert\,\pi_\theta(\cdot\mid c)\right)
+\mathcal H(p),
\label{eq:sequence-distillation}
\end{equation}
where $\mathcal H(p)$ is independent of $\theta$ \citep{kim2016sequence}.
Let $\widehat P_v(c,y)$ be the empirical distribution of stored pairs for
view $v$, with context marginal $\widehat P_v(c)=\sum_y\widehat P_v(c,y)$
and conditional teacher $q_v(y\mid c)=\widehat P_v(y\mid c)$. Conditioning
on the prompt and applying Equation~\ref{eq:sequence-distillation} gives
\begin{equation}
\begin{aligned}
\mathbb E_{(c,y)\sim\widehat P_v}[-\log\pi_\theta(y\mid c)]
&=\mathbb E_{c\sim\widehat P_v}\bigl[
D_{\mathrm{KL}}(q_v(\cdot\mid c)\Vert\pi_\theta(\cdot\mid c))\bigr]\\
&\quad+\mathbb E_{c\sim\widehat P_v}
\mathcal H(q_v(\cdot\mid c)).
\end{aligned}
\label{eq:empirical-nll-kl}
\end{equation}
For T, I, and R, $q_v$ is the corresponding conditional teacher under
Equation~\ref{eq:view-pairs}. Thus per-record NLL averaging recovers
$\mathcal L_T$, $\mathcal L_I$, and $\mathcal L_R$ up to constant entropies.
Individual targets have zero entropy; pooling records with a shared prompt
produces the conditional mixture in Equation~\ref{eq:empirical-nll-kl}.

\paragraph{Normalization and implementation.}
For a target $y$ of $|y|$ tokens, $y_j$ is token $j$ and $y_{<j}$ its
preceding prefix. The trainer uses the completion-length-normalized loss
\begin{equation}
\ell_\theta(y,c)=
-\frac{1}{|y|}\sum_{j=1}^{|y|}
\log\pi_\theta(y_j\mid c,y_{<j}),
\label{eq:length-normalized-nll}
\end{equation}
which normalizes short documents, final responses, and longer tool-call waves
by their completion lengths. For the stored point targets, this is the
sequence-level KL in Equation~\ref{eq:hard-target-kl} weighted by $1/|y|$.

The implemented objective samples retained records uniformly from
$\mathcal D_H^{\mathrm{rep}}$ and selects view $v$ with probability $\alpha_v$:

\begin{equation}
\mathcal L_{\mathrm{HPD}}^{\mathrm{impl}}
=\mathbb E_{(x_b^\star,d_b^H,\tilde a_b)
\sim\mathcal D_H^{\mathrm{rep}}}
\mathbb E_{v\sim\operatorname{Cat}(\alpha_T,\alpha_I,\alpha_R)}
\mathbb E_{y\sim p_b^v}
\left[\ell_\theta(y,c_b^v)\right].
\label{eq:implemented-objective}
\end{equation}
Each inner expectation evaluates the stored target: $\tilde a_b$ for T/I
and $d_b^H$ for R. A deterministic cycle realizes the long-run view weights
through sequential AdamW updates, each using 128 sequences from one view's
seeded permutation. One T/I/R round is one training step, giving default
weights $\alpha_T=\alpha_I=\alpha_R=1/3$ and 60k optimizer updates over 20k
training steps. All views update the same parameters; deployment uses only
$a_b\sim\pi_\theta^T(\cdot\mid x_b)$ under the native harness contract.

\end{document}

%% file: math_commands.tex
\usepackage{amsmath,amsfonts,bm}

\def\eqref#1{equation~\ref{#1}}

\def\1{\bm{1}}

\DeclareMathAlphabet{\mathsfit}{\encodingdefault}{\sfdefault}{m}{sl}
\SetMathAlphabet{\mathsfit}{bold}{\encodingdefault}{\sfdefault}{bx}{n}



%% file: hpd_overview.tex
\begingroup
\setlength{\fboxsep}{6pt}
\colorlet{hpdblue}{blue!45!black}
\colorlet{hpdgreen}{green!35!black}
\fcolorbox{hpdblue}{blue!3}{\begin{minipage}{0.94\linewidth}
\centering\small
\textbf{Offline teacher construction at boundary $b$}\\[4pt]
Reference prefix $\xrightarrow{\text{replay in }H}$
state $s_b^\star$ and native context $x_b^\star$\\[3pt]
Pre-action facts + reference ledger + action witness $a_b^\star$
$\xrightarrow{\textsc{AutoProbe}} d_b^H$\\[3pt]
Frozen model $\pi_{\theta_0}(\cdot\mid\operatorname{Inject}_H(x_b^\star,d_b^H))$
$\xrightarrow{K=2}$ candidate actions\\[3pt]
Harness verification $\longrightarrow$ nearest-reference selection
$\longrightarrow$ one hard target $\tilde a_b$\\[3pt]
\emph{The recorded action $a_b^\star$ advances replay; $\tilde a_b$ supervises training.}
\end{minipage}}\\[5pt]
$\Downarrow$\quad retained record $(x_b^\star,d_b^H,\tilde a_b)$\\[5pt]
\fcolorbox{hpdgreen}{green!3}{\begin{minipage}{0.94\linewidth}
\centering\small
\textbf{One shared student, three interleaved training views}\\[4pt]
\begin{tabular}{@{}lll@{}}
T: \replayaction & native context $x_b^\star$ & $\longrightarrow\tilde a_b$ \\
I: \injected & injected context $(x_b^\star,d_b^H)$ & $\longrightarrow\tilde a_b$ \\
R: \runtime & probe of native context $x_b^\star$ & $\longrightarrow d_b^H$
\end{tabular}\\[4pt]
$\mathcal L_{\mathrm{HPD}}=\alpha_T\mathcal L_T+
\alpha_I\mathcal L_I+\alpha_R\mathcal L_R$
\end{minipage}}\\[5pt]
$\Downarrow$\quad \textbf{Deployment:} native context $x_b$
$\longrightarrow\pi_\theta^T\longrightarrow a_b\longrightarrow H$\\[2pt]
{\small No control document or additional model call at deployment.}
\endgroup

%% file: hpd_bfcl_ood.tex
\begingroup
\usetikzlibrary{patterns}
\definecolor{bfclBaseline}{HTML}{AAB8C5}
\definecolor{bfclHPD}{HTML}{2369A1}
\definecolor{bfclHPDT}{HTML}{7AA9CD}
\definecolor{bfclHatch}{HTML}{163C5C}
\resizebox{\linewidth}{!}{%
\begin{tikzpicture}
\begin{groupplot}[
  group style={group size=3 by 1,horizontal sep=0.45cm,yticklabels at=edge left},
  scale only axis,width=0.27\linewidth,height=0.46\linewidth,
  xbar,/pgf/bar width=10pt,/pgf/bar shift=0pt,
  xmin=0,xmax=62,ymin=-0.6,ymax=7.6,y dir=reverse,
  xtick={0,20,40,60},ytick={0,1,2,3,4,5,6,7},
  yticklabels={Base,SFT,DPO,GRPO,OPSD,OPD,HPD-T,\textbf{HPD}},
  xlabel={Score (\%)},
  axis lines*=left,y axis line style={draw=none},
  axis line style={gray!40},tick style={draw=none},
  xmajorgrids,grid style={gray!20},clip=false,
  tick label style={font=\small},label style={font=\small},
  title style={font=\small},
  point meta=x,
  nodes near coords={\pgfmathprintnumber[fixed,precision=2,zerofill]{\pgfplotspointmeta}},
  nodes near coords style={anchor=west,xshift=2pt,font=\small,text=black},
  bfcl baseline/.style={fill=bfclBaseline,draw=none},
  bfcl hpd-t/.style={fill=bfclHPDT,draw=none},
  bfcl hpd/.style={fill=bfclHPD,draw=bfclHatch,
    postaction={pattern=north east lines,pattern color=bfclHatch},
    nodes near coords style={font=\small\bfseries}},
]
\nextgroupplot[title={(a) TS}]
\addplot[bfcl baseline] coordinates {
  (13.04,0) (0.00,1) (17.39,2) (8.70,3) (8.70,4) (17.39,5)
};
\addplot[bfcl hpd-t] coordinates {(13.04,6)};
\addplot[bfcl hpd] coordinates {(21.74,7)};
\nextgroupplot[title={(b) F1}]
\addplot[bfcl baseline] coordinates {
  (19.46,0) (15.75,1) (15.71,2) (18.85,3) (16.89,4) (21.50,5)
};
\addplot[bfcl hpd-t] coordinates {(26.12,6)};
\addplot[bfcl hpd] coordinates {(23.29,7)};
\nextgroupplot[title={(c) Step-SR}]
\addplot[bfcl baseline] coordinates {
  (42.03,0) (27.17,1) (41.67,2) (37.68,3) (44.06,4) (45.87,5)
};
\addplot[bfcl hpd-t] coordinates {(46.16,6)};
\addplot[bfcl hpd] coordinates {(48.12,7)};
\end{groupplot}
\end{tikzpicture}%
}
\endgroup

%% file: hpd_arxiv.bbl
\begin{thebibliography}{42}
\providecommand{\natexlab}[1]{#1}
\providecommand{\url}[1]{\texttt{#1}}
\expandafter\ifx\csname urlstyle\endcsname\relax
  \providecommand{\doi}[1]{doi: #1}\else
  \providecommand{\doi}{doi: \begingroup \urlstyle{rm}\Url}\fi

\bibitem[Agarwal et~al.(2024)Agarwal, Vieillard, Zhou,
  et~al.]{agarwal2024onpolicy}
Rishabh Agarwal, Nino Vieillard, Yongchao Zhou, et~al.
\newblock On-policy distillation of language models: Learning from
  self-generated mistakes.
\newblock In \emph{International Conference on Learning Representations}, 2024.
\newblock URL
  \url{https://proceedings.iclr.cc/paper_files/paper/2024/hash/5be69a584901a26c521c2b51e40a4c20-Abstract-Conference.html}.

\bibitem[Bengio et~al.(2015)Bengio, Vinyals, Jaitly, and
  Shazeer]{bengio2015scheduled}
Samy Bengio, Oriol Vinyals, Navdeep Jaitly, and Noam Shazeer.
\newblock Scheduled sampling for sequence prediction with recurrent neural
  networks.
\newblock In \emph{Advances in Neural Information Processing Systems},
  volume~28, 2015.
\newblock URL
  \url{https://proceedings.neurips.cc/paper_files/paper/2015/hash/e995f98d56967d946471af29d7bf99f1-Abstract.html}.

\bibitem[Chen et~al.(2021)Chen, Lu, Rajeswaran, et~al.]{chen2021decision}
Lili Chen, Kevin Lu, Aravind Rajeswaran, et~al.
\newblock Decision transformer: Reinforcement learning via sequence modeling.
\newblock In \emph{Advances in Neural Information Processing Systems},
  volume~34, 2021.
\newblock URL
  \url{https://proceedings.neurips.cc/paper_files/paper/2021/hash/7f489f642a0ddb10272b5c31057f0663-Abstract.html}.

\bibitem[Feng et~al.(2025)Feng, Xue, Liu, and An]{feng2025gigpo}
Lang Feng, Zhenghai Xue, Tingcong Liu, and Bo~An.
\newblock Group-in-group policy optimization for {LLM} agent training.
\newblock In \emph{Advances in Neural Information Processing Systems},
  volume~38, 2025.
\newblock URL
  \url{https://proceedings.neurips.cc/paper_files/paper/2025/hash/420c9f777c0b4f78d515e53cf74d58b2-Abstract-Conference.html}.

\bibitem[{Gemma Team} et~al.(2026)]{gemmateam2026gemma4}
{Gemma Team} et~al.
\newblock {Gemma 4} technical report.
\newblock arXiv preprint arXiv:2607.02770, 2026.
\newblock URL \url{https://arxiv.org/abs/2607.02770}.

\bibitem[Guo et~al.(2025)Guo, Yang, Zhang, et~al.]{deepseekai2025r1}
Daya Guo, Dejian Yang, Haowei Zhang, et~al.
\newblock {DeepSeek-R1} incentivizes reasoning in {LLMs} through reinforcement
  learning.
\newblock \emph{Nature}, 645\penalty0 (8081):\penalty0 633--638, 2025.
\newblock \doi{10.1038/s41586-025-09422-z}.
\newblock URL \url{https://www.nature.com/articles/s41586-025-09422-z}.

\bibitem[He et~al.(2026)He, Dai, He, Liu, Tang, Lu, Li, Ding, Mukherjee, Wang,
  Xing, Tang, and Dumoulin]{he2026trajectbench}
Pengfei He, Zhenwei Dai, Bing He, Hui Liu, Xianfeng Tang, Hanqing Lu, Juanhui
  Li, Jiayuan Ding, Subhabrata Mukherjee, Suhang Wang, Yue Xing, Jiliang Tang,
  and Benoit Dumoulin.
\newblock {TRAJECT-Bench}: A trajectory-aware benchmark for evaluating agentic
  tool use.
\newblock In \emph{International Conference on Learning Representations}, 2026.
\newblock URL \url{https://openreview.net/forum?id=TZWnWvsQ0X}.

\bibitem[Hinton et~al.(2015)Hinton, Vinyals, and Dean]{hinton2015distilling}
Geoffrey Hinton, Oriol Vinyals, and Jeff Dean.
\newblock Distilling the knowledge in a neural network.
\newblock arXiv preprint arXiv:1503.02531, 2015.
\newblock URL \url{https://arxiv.org/abs/1503.02531}.
\newblock Presented at the NIPS 2014 Deep Learning Workshop.

\bibitem[Hu et~al.(2022)Hu, Shen, Wallis, et~al.]{hu2022lora}
Edward~J. Hu, Yelong Shen, Phillip Wallis, et~al.
\newblock {LoRA}: Low-rank adaptation of large language models.
\newblock In \emph{International Conference on Learning Representations}, 2022.
\newblock URL \url{https://openreview.net/forum?id=nZeVKeeFYf9}.

\bibitem[Janner et~al.(2021)Janner, Li, and Levine]{janner2021trajectory}
Michael Janner, Qiyang Li, and Sergey Levine.
\newblock Offline reinforcement learning as one big sequence modeling problem.
\newblock In \emph{Advances in Neural Information Processing Systems},
  volume~34, 2021.
\newblock URL
  \url{https://proceedings.neurips.cc/paper_files/paper/2021/hash/099fe6b0b444c23836c4a5d07346082b-Abstract.html}.

\bibitem[Jimenez et~al.(2024)Jimenez, Yang, Wettig,
  et~al.]{jimenez2024swebench}
Carlos~E. Jimenez, John Yang, Alexander Wettig, et~al.
\newblock {SWE-bench}: Can language models resolve real-world {GitHub} issues?
\newblock In \emph{International Conference on Learning Representations}, 2024.
\newblock URL
  \url{https://proceedings.iclr.cc/paper_files/paper/2024/hash/edac78c3e300629acfe6cbe9ca88fb84-Abstract-Conference.html}.

\bibitem[Kim \& Rush(2016)Kim and Rush]{kim2016sequence}
Yoon Kim and Alexander~M. Rush.
\newblock Sequence-level knowledge distillation.
\newblock In \emph{Proceedings of the 2016 Conference on Empirical Methods in
  Natural Language Processing}, pp.\  1317--1327. Association for Computational
  Linguistics, 2016.
\newblock \doi{10.18653/v1/D16-1139}.
\newblock URL \url{https://aclanthology.org/D16-1139/}.

\bibitem[Lightman et~al.(2024)Lightman, Kosaraju, Burda,
  et~al.]{lightman2024verify}
Hunter Lightman, Vineet Kosaraju, Yura Burda, et~al.
\newblock Let's verify step by step.
\newblock In \emph{International Conference on Learning Representations}, 2024.
\newblock URL
  \url{https://proceedings.iclr.cc/paper_files/paper/2024/hash/aca97732e30bcf1303bc22ac3924fd16-Abstract-Conference.html}.

\bibitem[Liu et~al.(2026)Liu, Dong, Lu, et~al.]{liu2026gdpo}
Shih-Yang Liu, Xin Dong, Ximing Lu, et~al.
\newblock {GDPO}: Group reward-decoupled normalization policy optimization for
  multi-reward {RL} optimization.
\newblock In \emph{Proceedings of the 43rd International Conference on Machine
  Learning}, 2026.
\newblock URL \url{https://openreview.net/forum?id=Y1rHxA58If}.

\bibitem[Liu et~al.(2024)Liu, Yu, Zhang, et~al.]{liu2024agentbench}
Xiao Liu, Hao Yu, Hanchen Zhang, et~al.
\newblock {AgentBench}: Evaluating {LLMs} as agents.
\newblock In \emph{International Conference on Learning Representations}, 2024.
\newblock URL
  \url{https://proceedings.iclr.cc/paper_files/paper/2024/hash/e9df36b21ff4ee211a8b71ee8b7e9f57-Abstract-Conference.html}.

\bibitem[Lopez-Paz et~al.(2016)Lopez-Paz, Bottou, Sch\"olkopf, and
  Vapnik]{lopezpaz2016unifying}
David Lopez-Paz, L\'eon Bottou, Bernhard Sch\"olkopf, and Vladimir Vapnik.
\newblock Unifying distillation and privileged information.
\newblock In \emph{International Conference on Learning Representations}, 2016.
\newblock URL \url{https://arxiv.org/abs/1511.03643}.

\bibitem[Lu et~al.(2025)Lu, Holleis, Zhang, et~al.]{lu2025toolsandbox}
Jiarui Lu, Thomas Holleis, Yizhe Zhang, et~al.
\newblock {ToolSandbox}: A stateful, conversational, interactive evaluation
  benchmark for {LLM} tool use capabilities.
\newblock In \emph{Findings of the Association for Computational Linguistics:
  NAACL 2025}, pp.\  1160--1183. Association for Computational Linguistics,
  2025.
\newblock \doi{10.18653/v1/2025.findings-naacl.65}.
\newblock URL \url{https://aclanthology.org/2025.findings-naacl.65/}.

\bibitem[Ma et~al.(2024)Ma, Zhang, Zhu, et~al.]{ma2024agentboard}
Chang Ma, Junlei Zhang, Zhihao Zhu, et~al.
\newblock {AgentBoard}: An analytical evaluation board of multi-turn {LLM}
  agents.
\newblock In \emph{Advances in Neural Information Processing Systems},
  volume~37, 2024.
\newblock URL
  \url{https://proceedings.neurips.cc/paper_files/paper/2024/hash/877b40688e330a0e2a3fc24084208dfa-Abstract-Datasets_and_Benchmarks_Track.html}.

\bibitem[Mialon et~al.(2024)Mialon, Fourrier, Swift, et~al.]{mialon2024gaia}
Gr\'egoire Mialon, Cl\'ementine Fourrier, Craig Swift, et~al.
\newblock {GAIA}: A benchmark for general {AI} assistants.
\newblock In \emph{International Conference on Learning Representations}, 2024.
\newblock URL
  \url{https://proceedings.iclr.cc/paper_files/paper/2024/hash/25ae35b5b1738d80f1f03a8713e405ec-Abstract-Conference.html}.

\bibitem[Ouyang et~al.(2022)Ouyang, Wu, Jiang, et~al.]{ouyang2022training}
Long Ouyang, Jeffrey Wu, Xu~Jiang, et~al.
\newblock Training language models to follow instructions with human feedback.
\newblock In \emph{Advances in Neural Information Processing Systems},
  volume~35, 2022.
\newblock URL
  \url{https://proceedings.neurips.cc/paper_files/paper/2022/hash/b1efde53be364a73914f58805a001731-Abstract-Conference.html}.

\bibitem[Patil et~al.(2025)Patil, Mao, Yan, Ji, Suresh, Stoica, and
  Gonzalez]{patil2025bfcl}
Shishir~G Patil, Huanzhi Mao, Fanjia Yan, Charlie Cheng-Jie Ji, Vishnu Suresh,
  Ion Stoica, and Joseph~E. Gonzalez.
\newblock The berkeley function calling leaderboard ({BFCL}): From tool use to
  agentic evaluation of large language models.
\newblock In \emph{Proceedings of the 42nd International Conference on Machine
  Learning}, volume 267 of \emph{Proceedings of Machine Learning Research},
  pp.\  48371--48392. PMLR, 2025.
\newblock URL \url{https://proceedings.mlr.press/v267/patil25a.html}.

\bibitem[Penaloza et~al.(2026)Penaloza, Vattikonda, Gontier, Lacoste, Charlin,
  and Caccia]{penaloza2026privileged}
Emiliano Penaloza, Dheeraj Vattikonda, Nicolas Gontier, Alexandre Lacoste,
  Laurent Charlin, and Massimo Caccia.
\newblock Privileged information distillation for language models.
\newblock In \emph{Proceedings of the 43rd International Conference on Machine
  Learning}, 2026.
\newblock URL \url{https://arxiv.org/abs/2602.04942}.

\bibitem[Qin et~al.(2024)Qin, Liang, Ye, et~al.]{qin2024toolllm}
Yujia Qin, Shihao Liang, Yining Ye, et~al.
\newblock {ToolLLM}: Facilitating large language models to master 16000+
  real-world {APIs}.
\newblock In \emph{International Conference on Learning Representations}, 2024.
\newblock URL
  \url{https://proceedings.iclr.cc/paper_files/paper/2024/hash/28e50ee5b72e90b50e7196fde8ea260e-Abstract-Conference.html}.

\bibitem[Rafailov et~al.(2023)Rafailov, Sharma, Mitchell,
  et~al.]{rafailov2023direct}
Rafael Rafailov, Archit Sharma, Eric Mitchell, et~al.
\newblock Direct preference optimization: Your language model is secretly a
  reward model.
\newblock In \emph{Advances in Neural Information Processing Systems},
  volume~36, 2023.
\newblock URL
  \url{https://proceedings.neurips.cc/paper_files/paper/2023/hash/a85b405ed65c6477a4fe8302b5e06ce7-Abstract-Conference.html}.

\bibitem[Ross et~al.(2011)Ross, Gordon, and Bagnell]{ross2011reduction}
St\'ephane Ross, Geoffrey~J. Gordon, and J.~Andrew Bagnell.
\newblock A reduction of imitation learning and structured prediction to
  no-regret online learning.
\newblock In \emph{Proceedings of the Fourteenth International Conference on
  Artificial Intelligence and Statistics}, volume~15, pp.\  627--635. PMLR,
  2011.
\newblock URL \url{https://proceedings.mlr.press/v15/ross11a.html}.

\bibitem[Schick et~al.(2023)Schick, Dwivedi-Yu, Dess\`i,
  et~al.]{schick2023toolformer}
Timo Schick, Jane Dwivedi-Yu, Roberto Dess\`i, et~al.
\newblock {Toolformer}: Language models can teach themselves to use tools.
\newblock In \emph{Advances in Neural Information Processing Systems},
  volume~36, 2023.
\newblock URL
  \url{https://proceedings.neurips.cc/paper_files/paper/2023/hash/d842425e4bf79ba039352da0f658a906-Abstract-Conference.html}.

\bibitem[Shao et~al.(2024)Shao, Wang, Zhu, et~al.]{shao2024deepseekmath}
Zhihong Shao, Peiyi Wang, Qihao Zhu, et~al.
\newblock {DeepSeekMath}: Pushing the limits of mathematical reasoning in open
  language models.
\newblock arXiv preprint arXiv:2402.03300, 2024.
\newblock URL \url{https://arxiv.org/abs/2402.03300}.

\bibitem[Shridhar et~al.(2021)Shridhar, Yuan, C\^ot\'e,
  et~al.]{shridhar2021alfworld}
Mohit Shridhar, Xingdi Yuan, Marc-Alexandre C\^ot\'e, et~al.
\newblock {ALFWorld}: Aligning text and embodied environments for interactive
  learning.
\newblock In \emph{International Conference on Learning Representations}, 2021.
\newblock URL \url{https://openreview.net/forum?id=0IOX0YcCdTn}.

\bibitem[Uesato et~al.(2022)Uesato, Kushman, Kumar, et~al.]{uesato2022solving}
Jonathan Uesato, Nate Kushman, Ramana Kumar, et~al.
\newblock Solving math word problems with process- and outcome-based feedback.
\newblock arXiv preprint arXiv:2211.14275, 2022.
\newblock URL \url{https://arxiv.org/abs/2211.14275}.
\newblock Presented at the NeurIPS 2022 MATH-AI Workshop.

\bibitem[Vapnik \& Vashist(2009)Vapnik and Vashist]{vapnik2009new}
Vladimir Vapnik and Akshay Vashist.
\newblock A new learning paradigm: Learning using privileged information.
\newblock \emph{Neural Networks}, 22\penalty0 (5--6):\penalty0 544--557, 2009.
\newblock \doi{10.1016/j.neunet.2009.06.042}.

\bibitem[Wang et~al.(2025)Wang, Li, Song, et~al.]{wang2025openhands}
Xingyao Wang, Boxuan Li, Yufan Song, et~al.
\newblock {OpenHands}: An open platform for {AI} software developers as
  generalist agents.
\newblock In \emph{International Conference on Learning Representations}, 2025.
\newblock URL
  \url{https://proceedings.iclr.cc/paper_files/paper/2025/hash/a4b6ad6b48850c0c331d1259fc66a69c-Abstract-Conference.html}.

\bibitem[Yang et~al.(2025)Yang, Li, Yang, et~al.]{yang2025qwen3}
An~Yang, Anfeng Li, Baosong Yang, et~al.
\newblock {Qwen3} technical report.
\newblock arXiv preprint arXiv:2505.09388, 2025.
\newblock URL \url{https://arxiv.org/abs/2505.09388}.

\bibitem[Yang et~al.(2024)Yang, Jimenez, Wettig, et~al.]{yang2024sweagent}
John Yang, Carlos~E. Jimenez, Alexander Wettig, et~al.
\newblock {SWE-agent}: Agent-computer interfaces enable automated software
  engineering.
\newblock In \emph{Advances in Neural Information Processing Systems},
  volume~37, 2024.
\newblock URL
  \url{https://proceedings.neurips.cc/paper_files/paper/2024/hash/5a7c947568c1b1328ccc5230172e1e7c-Abstract-Conference.html}.

\bibitem[Yang et~al.(2023)Yang, Song, Li, et~al.]{yang2023gpt4tools}
Rui Yang, Lin Song, Yanwei Li, et~al.
\newblock {GPT4Tools}: Teaching large language model to use tools via
  self-instruction.
\newblock In \emph{Advances in Neural Information Processing Systems},
  volume~36, 2023.
\newblock URL
  \url{https://proceedings.neurips.cc/paper_files/paper/2023/hash/e393677793767624f2821cec8bdd02f1-Abstract-Conference.html}.

\bibitem[Yao et~al.(2022)Yao, Chen, Yang, and Narasimhan]{yao2022webshop}
Shunyu Yao, Howard Chen, John Yang, and Karthik Narasimhan.
\newblock {WebShop}: Towards scalable real-world web interaction with grounded
  language agents.
\newblock In \emph{Advances in Neural Information Processing Systems},
  volume~35, 2022.
\newblock URL
  \url{https://proceedings.neurips.cc/paper_files/paper/2022/hash/82ad13ec01f9fe44c01cb91814fd7b8c-Abstract-Conference.html}.

\bibitem[Yao et~al.(2023)Yao, Zhao, Yu, et~al.]{yao2023react}
Shunyu Yao, Jeffrey Zhao, Dian Yu, et~al.
\newblock {ReAct}: Synergizing reasoning and acting in language models.
\newblock In \emph{International Conference on Learning Representations}, 2023.
\newblock URL \url{https://openreview.net/forum?id=WE_vluYUL-X}.

\bibitem[Yao et~al.(2025)Yao, Shinn, Razavi, and Narasimhan]{yao2025taubench}
Shunyu Yao, Noah Shinn, Pedram Razavi, and Karthik Narasimhan.
\newblock {$\tau$-bench}: A benchmark for tool-agent-user interaction in
  real-world domains.
\newblock In \emph{International Conference on Learning Representations}, 2025.
\newblock URL
  \url{https://proceedings.iclr.cc/paper_files/paper/2025/hash/1b126cc38b8638e07bef37e7b2bb72bf-Abstract-Conference.html}.

\bibitem[Zhao et~al.(2026{\natexlab{a}})Zhao, Xie, Liu, Huang, Pang, Chen, and
  Grover]{zhao2026selfdistilled}
Siyan Zhao, Zhihui Xie, Mengchen Liu, Jing Huang, Guan Pang, Feiyu Chen, and
  Aditya Grover.
\newblock Self-distilled reasoner: On-policy self-distillation for large
  language models.
\newblock In \emph{Proceedings of the 43rd International Conference on Machine
  Learning}, 2026{\natexlab{a}}.
\newblock URL \url{https://openreview.net/forum?id=Jpxfof0EaS}.

\bibitem[Zhao et~al.(2026{\natexlab{b}})Zhao, Ma, and
  Zhang]{zhao2026trainingharnesses}
Zhengyang Zhao, Lu~Ma, and Wentao Zhang.
\newblock Training with harnesses: On-policy harness self-distillation for
  complex reasoning.
\newblock arXiv preprint arXiv:2605.08741, 2026{\natexlab{b}}.
\newblock URL \url{https://arxiv.org/abs/2605.08741}.

\bibitem[Zheng et~al.(2025)Zheng, Liu, Li, et~al.]{zheng2025gspo}
Chujie Zheng, Shixuan Liu, Mingze Li, et~al.
\newblock Group sequence policy optimization.
\newblock arXiv preprint arXiv:2507.18071, 2025.
\newblock URL \url{https://arxiv.org/abs/2507.18071}.

\bibitem[Zhou et~al.(2024)Zhou, Xu, Zhu, et~al.]{zhou2024webarena}
Shuyan Zhou, Frank~F. Xu, Hao Zhu, et~al.
\newblock {WebArena}: A realistic web environment for building autonomous
  agents.
\newblock In \emph{International Conference on Learning Representations}, 2024.
\newblock URL
  \url{https://proceedings.iclr.cc/paper_files/paper/2024/hash/4410c0711e9154a7a2d26f9b3816d1ef-Abstract-Conference.html}.

\bibitem[Zhou et~al.(2026)Zhou, Zheng, Li, Peng, Xu, and Chen]{zhou2026turnopd}
Yuhang Zhou, Kai Zheng, Haoling Li, Dengyun Peng, Can Xu, and Jingjing Chen.
\newblock {TurnOPD}: Making on-policy distillation turn-aware for efficient
  long-horizon agent training.
\newblock arXiv preprint arXiv:2607.05804, 2026.
\newblock URL \url{https://arxiv.org/abs/2607.05804}.

\end{thebibliography}
